\documentclass[10pt,twocolumn,letterpaper]{article}

\usepackage{paper}              % To produce the CAMERA-READY version

\definecolor{myblue}{rgb}{0.21,0.49,0.74}
\usepackage[pagebackref,breaklinks,colorlinks,allcolors=myblue]{hyperref}
\usepackage{multirow}

\usepackage[ruled,vlined,linesnumbered]{algorithm2e}
\title{The Devil is in the Distributions: Explicit Modeling of Scene Content is Key in Zero-Shot Video Captioning}

\author{Mingkai Tian\textsuperscript{1} \quad Guorong Li\textsuperscript{1} \quad Yuankai Qi\textsuperscript{2} \quad  Amin Beheshti\textsuperscript{2} \\
 Javen Qinfeng Shi\textsuperscript{3} \quad  Anton van den Hengel\textsuperscript{3} \quad Qingming Huang\textsuperscript{1}\\
\textsuperscript{1}School of Computer Science and Technology, University of Chinese Academy of Sciences \\
\textsuperscript{2}Macquarie University \quad \textsuperscript{3}Australian Institute for Machine Learning, The University of Adelaide \\
{\tt\small liguorong@ucas.ac.cn}}

\begin{document}
\maketitle
\begin{abstract}
% Video captioning, the task of generating textual descriptions for video content, has traditionally relied on supervised methods that require large-scale annotated video-text pairs, limiting their scalability in real-world applications. To address this, zero-shot video captioning has emerged as a promising direction, aiming to generate high-quality captions without the need for paired training data. 
Zero-shot video captioning requires that a model generate high-quality captions without  human-annotated video-text pairs for training.
%
%A While existing methods have made significant progress by leveraging textual semantic units as prefix prompts, they often suffer from limitations in capturing comprehensive video content. 
%original A%  The state-of-the-art approach to the problem leverages pre-trained models, and particularly uses an LLM to generate potential captions that are guided using CLIP.
State-of-the-art approaches to the problem leverage CLIP to extract visual-relevant textual prompts to guide language models in generating captions.
%A These methods typically employ single-granularity prompts (\eg, noun phrases, entire sentences), leading to either a lack of interaction between objects or a dilution of fine-grained details. 
These methods tend to focus on one key aspect of the scene and build a caption that ignores the rest of the visual input.
%Furthermore, the common practice of selecting the top-\(K\) most relevant semantic units often leads to a dominance of semantically similar elements (\eg, multiple noun phrases describing the same entity), thereby compromising prompt diversity and comprehensiveness.
% and, consequently, impacting the accuracy of the generated captions.
%
To address this issue, and generate more accurate and complete captions, we propose a novel progressive multi-granularity textual prompting strategy for zero-shot video captioning. Our approach constructs three distinct memory banks, encompassing noun phrases, scene graphs of noun phrases, 
and entire sentences. 
% By extracting progressively richer information from these banks, our model provides a more complete semantic context for the captioning process. 
Moreover, we introduce a category-aware retrieval mechanism 
%A coupled with top-\textit{p} post-processing, a strategy that promotes prompt diversity and reduces redundancy while ensuring visual relevance.
that models the distribution of natural language surrounding the specific topics in question.
% , moving beyond traditional top-\textit{K} selection.
%
Extensive experiments demonstrate the effectiveness of our method with 5.7\%, 16.2\%, and 3.4\% improvements in terms of the main metric CIDEr on MSR-VTT, MSVD, and VATEX benchmarks compared to existing state-of-the-art.
\end{abstract}    
\section{Introduction}
\label{sec:intro}

% Distribution is the meaning of belonging to which category?

% Video captioning, the task of generating precise descriptions for video content, serves as a pivotal bridge between computer vision and natural language processing. It holds broad application prospects in domains such as video retrieval, accessibility for the visually impaired, and human-computer interaction. With the proliferation of short-form video platforms and live streaming, the demand for automated video captioning has surged, driving rapid advancements in this research area.

% 
%A Video captioning, the task of generating textual descriptions for video content, serves as a critical bridge between computer vision and natural language processing. It offers substantial value in applications such as video retrieval, accessibility for the visually impaired, and human-computer interaction. 
The fact that video captioning remains a challenge, despite significant effort, reflects the inherent complexity of understanding video.  Part of the specific problem with captioning is the vast difference between video and natural language as data forms.  Video represents a voluminous stream of continuous pixel measurements, whereas natural language is a sequence of discrete symbols with a peculiar structure. Methods that directly model associations between modalities are vulnerable to missing the structure in either.  This is visible in the fact that they are susceptible to generating captions that focus on a single scene element. We propose an approach that models the structure of scenes, and the specific language that describes them, at multiple scales in the hope that the resulting model might not be easily distracted.

% The rise of short-form video platforms and live streaming has fueled an urgent demand for automated video captioning, accelerating progress in this field.
% Traditional supervised video captioning methods~\cite{DBLP:conf/cvpr/swinbert, DBLP:journals/pr/tmk_paper} rely on large-scale, manually annotated video-text pairs for training, typically employing an encoder-decoder architecture.
% The encoder leverages pre-trained 
% % 2D or 3D 
% convolutional neural networks (\eg ResNet~\cite{DBLP:conf/cvpr/ResNet}, I3D~\cite{DBLP:conf/cvpr/I3D}, C3D~\cite{DBLP:conf/cvpr/C3D}) 
% % to extract visual features
% , while the decoder generates descriptions using LSTM~\cite{DBLP:journals/neco/HochreiterS97} or Transformer-based models~\cite{DBLP:conf/cvpr/Evcap, DBLP:conf/cvpr/M²_Transformer}. 
% Although they have achieved remarkable performance, their dependence on human-labeled data constrains their scalability in real-world applications.
Traditional supervised video captioning methods~\cite{DBLP:conf/cvpr/swinbert, DBLP:journals/pr/tmk_paper, DBLP:conf/cvpr/SAAT, DBLP:conf/iccv/POS-CG} utilize an encoder-decoder architecture trained on large-scale, manually annotated video-text pairs.
The encoder leverages pre-trained 
% 2D or 3D
convolutional neural networks (\eg, ResNet~\cite{DBLP:conf/cvpr/ResNet}, I3D~\cite{DBLP:conf/cvpr/I3D}, C3D~\cite{DBLP:conf/cvpr/C3D}), while the decoder uses LSTMs~\cite{DBLP:journals/neco/HochreiterS97} or Transformers~\cite{DBLP:conf/cvpr/Evcap, DBLP:conf/cvpr/M²_Transformer}.
Despite achieving remarkable performance, their reliance on human-labeled data constrains real-world scalability.
% Annotating large-scale video datasets is not only time-consuming and labor-intensive but also challenging to cover diverse domains and linguistic contexts.

% Traditional supervised video captioning methods~\cite{DBLP:conf/cvpr/swinbert, DBLP:journals/pr/tmk_paper} depend on large-scale, manually annotated video-text pairs for training and commonly use an encoder-decoder architecture. While they have demonstrated impressive performance, their reliance on human annotations constrains scalability in practical applications.

\begin{figure}[t]
  \centering
   \includegraphics[width=1\linewidth]{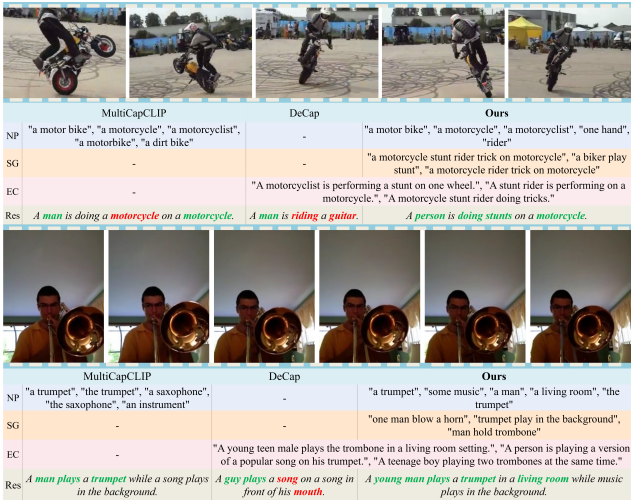}

   \caption{Current zero-shot captioning methods are easily distracted. NP, SG, and EC denote noun phrase, scene graph 
   % (containing noun phrases wherever possible)
   (triplets are displayed as concatenated strings), and entire caption prompt, respectively. 
   % \colorbox[RGB]{219,238,244}{\textcolor{black}{Res}} indicates the generated captions, 
   ``Res'' indicates the generated captions, with \textbf{\textit{\textcolor[RGB]{0,176,80}{correct}}} and \textbf{\textit{\textcolor[RGB]{254,0,0}{incorrect}}} words highlighted. 
   %
   %T In the top video example, MultiCapCLIP~\cite{DBLP:conf/acl/MultiCapCLIP} fails to capture the rider-motorcycle interaction because the prior does not model the distribution of object-object interactions effectively. 
   In the top example, MultiCapCLIP~\cite{DBLP:conf/acl/MultiCapCLIP} fails to capture the rider-motorcycle interaction as the prior does not model the distribution of subject-object interactions effectively.
   In the bottom example, MultiCapCLIP's top-\textit{K} retrieval strategy produces repetitive similar noun phrases and lacks person and environment information,
   % In the bottom example, the top-\textit{K} retrieval strategy of MultiCapCLIP leads to repetitive similar noun phrases 
   because it fails to model either the structure of the scene or of natural language. 
   DeCap~\cite{DBLP:conf/iclr/DeCap} struggles to fully understand the video details due to its coarse-grained prompt of global caption embedding. 
   By contrast, 
   % Conversely
   our method generates more accurate and comprehensive descriptions.}
   \label{fig:intro}
\end{figure}

To address these limitations, zero-shot video captioning has emerged as a promising direction, aiming to generate high-quality descriptions without relying on video-text pairs for training. Existing methods can be broadly categorized into two types: training-free approaches and those trained solely on text corpora.
Training-free methods leverage pre-trained vision-language models (\eg, CLIP~\cite{DBLP:conf/icml/CLIP}) to guide pre-trained language models (\eg, GPT-2~\cite{GPT_2}, BERT~\cite{DBLP:conf/naacl/BERT}) in generating text during inference. For instance, Tewel \etal~\cite{DBLP:conf/bmvc/ZeroCap_video} employ randomly initialized pseudo-tokens and prefix prompts (\eg, ``Video of'') to assist GPT-2 in generating new tokens. After generating a complete sentence, the model updates the pseudo-tokens based on the gradient derived from the CLIP cross-modal similarity.
% between the generated tokens and video frames, initiating the next round of sentence generation. 
% Similarly, other zero-shot image captioning methods like ZeroCap~\cite{DBLP:conf/cvpr/ZeroCap} and ConZIC~\cite{DBLP:conf/cvpr/ConZIC} update the language model's KV-cache or sample new tokens during inference using CLIP's cross-modal similarity. 
However, introducing visual supervision  after token generation can lead to the language model's priors dominating the captioning process, resulting in hallucinations unrelated to the video content. 
% Moreover, an ideal caption often requires multiple iterations of refinement.

The other line of work involves training a text decoder on pure text corpora. Textual units (\eg, nouns, noun phrases, complete sentences) are extracted from the training corpus to form various memory banks. During training, embeddings of these textual units
% , either retrieved or parsed from captions, 
serve as prefix prompts for the text decoder to reconstruct the original caption. At inference time, visual features are used to retrieve relevant textual semantic units from the memory bank via CLIP similarity, which are then fed into the text decoder. For example, MultiCapCLIP~\cite{DBLP:conf/acl/MultiCapCLIP} constructs a memory bank of noun phrases and retrieves top-\textit{K} elements as prefix prompts. DeCap~\cite{DBLP:conf/iclr/DeCap}, on the other hand, builds a memory bank of entire captions and uses a single token of global embedding as the prefix prompt. Other text-only trained zero-shot image captioning methods, such as MeaCap~\cite{DBLP:conf/cvpr/MeaCap}, create a memory bank of complete captions as well, retrieve the top-\textit{K} relevant descriptions, and parse key entities to input into a pre-trained language model. 
% Similarly, ViECap~\cite{DBLP:conf/iccv/ViECap} utilizes COCO and VGOI vocabularies as memory banks retrieving top-\textit{K} categories as hard prompts and using image embeddings as soft prompts during inference.
% Similarly, ViECap~\cite{DBLP:conf/iccv/ViECap} utilizes COCO and VGOI vocabularies as memory banks, employing noun embeddings and sentence token embeddings as hard and soft prompts during training, and retrieving top-K categories as hard prompts and using image embeddings as soft prompts during inference.

Despite the progress, the above-mentioned methods still face limitations in their prefix prompt construction and retrieval strategies. 
They often rely on single-granularity textual units, such as nouns, noun phrases, or complete sentences, as prompts, failing to fully exploit multi-granularity textual units to provide rich information for the language model. 
While noun phrases offer more attribute information than simple nouns, they lack interaction information between entities. In addition, the adopted global embeddings of complete sentences of existing methods may dilute fine-grained details. 
As illustrated in the top video example of~\cref{fig:intro}, MultiCapCLIP fails to accurately capture the rider's stunt action providing only noun phrase prompts, while DeCap fails to identify the ``motorcycle''. 
% , resulting in insufficient input for the language model. 
Furthermore, top-\textit{K} retrieval strategies, which simply select the \textit{K} most similar elements, tend to retrieve semantically repetitive textual units, reducing the diversity of the prompts and the accuracy of the generated captions. 
We provide such an example in ~\cref{fig:intro} bottom, where MultiCapCLIP repetitively prompts with musical instrument phrases but neglects information about the person and the environment.

To address these challenges, we propose a novel progressive multi-granularity textual prompting strategy. 
We construct three distinct memory banks comprising noun phrases, scene graphs that incorporate noun phrases, and entire sentences, ensuring the text decoder receives comprehensive semantic cues. Existing  parsers~\cite{DBLP:conf/acl/FACTUAL} typically extract scene graphs with noun-only nodes, while we propose a text-similarity-based approach to enhance initial scene graphs by incorporating noun phrases with additional attributes wherever possible. 
We further develop a category-aware retrieval mechanism with top-\textit{p} filtering for noun phrases and scene graphs, ensuring both diversity and visual relevance.
~\cref{fig:intro} shows the captions generated by our method.
% A category-aware retrieval mechanism with top-\textit{p} post-processing for noun phrases and scene graphs is also developed, ensuring visual relevance and diversity simultaneously.
% Furthermore, we introduce a category-aware retrieval mechanism with Top-P post-processing for both noun phrases and scene graphs, enhancing the prompt's comprehensiveness while ensuring its relevance to the visual features.

The main contributions of our work are summarized as follows:
\begin{itemize}
    \item We propose a progressive multi-granularity textual prompting strategy, providing the language model with comprehensive semantic information across varying levels of abstraction.
    \item We introduce a category-aware retrieval method with top-\textit{p} post-processing for semantic units, enhancing the diversity and relevance of the prompts 
    % fed into the language model
    during inference.
    % , thereby enhancing the quality of the generated descriptions.
    \item 
    % Compared to all previous methods~\cite{DBLP:conf/bmvc/ZeroCap_video, DBLP:conf/iclr/DeCap, DBLP:conf/acl/MultiCapCLIP}, which were only shallowly tested on video captioning datasets, 
    % We conduct in-domain and cross-domain experiments on three widely used video captioning datasets, 
    % and perform the first  experiments on two of them, 
    % establishing new SoTA performance across all metrics. 
    Extensive experiments on the MSR-VTT, MSVD, and VATEX benchmarks demonstrate the effectiveness of our method with 5.7\%, 16.2\%, and 3.4\% CIDEr improvements over state-of-the-art methods.
    % Extensive experiments demonstrate that our method achieves improvements of  in the main metric CIDEr on the MSR-VTT, MSVD, and VATEX benchmarks
    % Additionally, we validate the effectiveness and scalability of our approach through extensive ablation studies.
\end{itemize}

\section{Related Work}
\label{sec:related_work}

Zero-shot visual captioning research splits into two main categories. The first uses training-free methods, leveraging pre-trained language models to generate captions during inference. These methods either optimize internal contexts~\cite{DBLP:conf/cvpr/ZeroCap, DBLP:conf/bmvc/ZeroCap_video} or design sampling strategies to align tokens with visual inputs~\cite{DBLP:conf/cvpr/ConZIC, DBLP:journals/corr/MAGIC}. The second trains models on text-only corpora, reconstructing captions from semantic units like nouns or noun phrases~\cite{DBLP:conf/iclr/DeCap, DBLP:conf/iccv/ViECap, DBLP:conf/emnlp/CapDec, DBLP:conf/cvpr/MeaCap, DBLP:conf/acl/MultiCapCLIP, DBLP:journals/ijon/EntroCap}, mapping visual features to textual prompts for inference.
%\paragraph{Frozen vs. Trainable Language Models} 
\subsection{Frozen vs. Trainable Language Models}
Training-free and text-only training approaches both rely on pre-trained language models (\eg, BERT~\cite{DBLP:conf/naacl/BERT}, GPT-2~\cite{GPT_2}) and vision-language models (\eg, CLIP~\cite{DBLP:conf/icml/CLIP}). Training-free methods keep these models frozen. ZeroCap~\cite{DBLP:conf/cvpr/ZeroCap} and its variant for zero-shot video captioning~\cite{DBLP:conf/bmvc/ZeroCap_video} employ GPT-2 to iteratively predict new textual tokens, with cross-modal similarity calculation using CLIP after each token is generated. Subsequently, the two methods use gradient descent to respectively optimize the \textit{key}-\textit{value} cache and the prefix pseudo-tokens. ConZIC~\cite{DBLP:conf/cvpr/ConZIC} utilizes a pre-trained BERT-Base with bidirectional attention for Gibbs sampling. It further incorporates cross-modal matching scores from CLIP to determine each token finally.
% and control signals from models like SentiWordNet 
In contrast, text-only training approaches fine-tune the weights of a pre-trained language model like GPT-2 (\eg, CapDec~\cite{DBLP:conf/emnlp/CapDec}, ViECap~\cite{DBLP:conf/iccv/ViECap}, EntroCap~\cite{DBLP:journals/ijon/EntroCap}) or CBART~\cite{DBLP:conf/emnlp/CBART} (\eg, MeaCap~\cite{DBLP:conf/cvpr/MeaCap}), or train transformers from scratch (\eg, DeCap~\cite{DBLP:conf/iclr/DeCap}, MultiCapCLIP~\cite{DBLP:conf/acl/MultiCapCLIP}). Our approach trains a lightweight randomly initialized Transformer, demonstrating strong adaptability and favorable performance.
\subsection{Textual Memory Bank}

Text-only training methods for zero-shot visual captioning often employ a textual memory bank for efficient storage and rich semantics. ViECap~\cite{DBLP:conf/iccv/ViECap} and EntroCap~\cite{DBLP:journals/ijon/EntroCap} utilize a memory bank of object class names from Visual Genome~\cite{DBLP:journals/ijcv/Visual_Genome}. They concatenate embeddings of parsed nouns with global caption embeddings during training, and retrieve relevant nouns using CLIP during inference. DeCap~\cite{DBLP:conf/iclr/DeCap} and MeaCap~\cite{DBLP:conf/cvpr/MeaCap} build memory banks with all training captions, employing sentence-level and core noun embeddings as prefix prompts, respectively. MultiCapCLIP's~\cite{DBLP:conf/acl/MultiCapCLIP} memory bank is composed of the 1000 most frequent noun phrases parsed from training captions, allowing the decoder to generate sentences from concept prompts.
However, existing methods have not fully explored the potential of textual semantic units as prompts, as nouns and noun phrases often lack action-related information, while global sentence embeddings tend to dilute finer details, leading to incomplete information for the language decoder. To address this, we construct three separate memory banks composed of noun phrases, scene graphs incorporating noun phrases, and entire sentences, applying a progressive multi-granularity prompting strategy to ensure comprehensive information for the language model, leading to excellent experimental performance.

\section{Method}

\begin{figure*}
    \centering
    \includegraphics[width=1\linewidth]{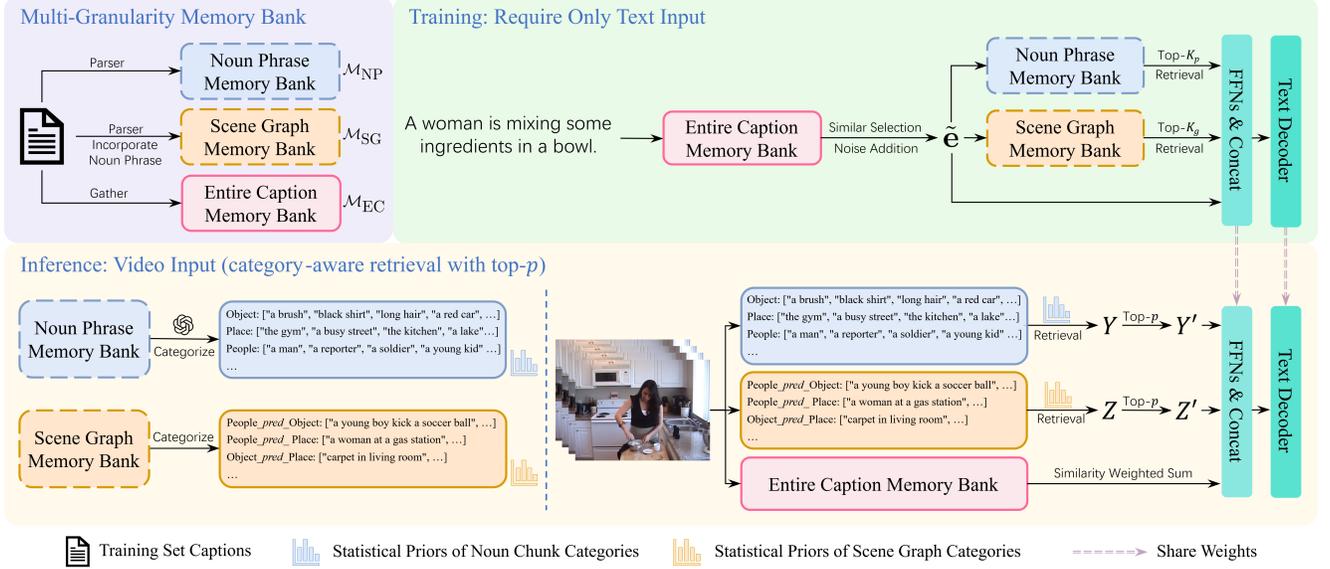}
    \caption{
    We construct noun phrase memory bank \(\mathcal{M}_{\text{NP}}\) and scene graph memory bank \(\mathcal{M}_{\text{SG}}\) by parsing training captions, selecting high-frequency elements, and enhancing scene graphs with noun phrases to include more attribute information. The entire caption memory bank \(\mathcal{M}_{\text{EC}}\) contains all training captions. 
    During training, following MultiCapCLIP~\cite{DBLP:conf/acl/MultiCapCLIP}, we retrieve top-\textit{K} elements from memory banks using perturbed embedding \(\tilde{\mathbf{e}}\) and train a text decoder to reconstruct the original text. 
    During inference, we first classify \(\mathcal{M}_{\text{NP}}\) with GPT-4~\cite{gpt4} and \(\mathcal{M}_{\text{SG}}\) based on noun phrase categories, compute statistical priors, retrieve a diverse set of relevant noun phrase and scene graph elements using CLIP embeddings with top-\textit{p} filtering and generate a weighted embedding from 
 \(\mathcal{M}_{\text{EC}}\) using softmax similarity scores between video and caption features. Three types of prompt are transformed by respective FFNs and concatenated to generate the final caption.
    % We show scene graph with concat sub, pred and obj of a triplet for simplity.
    }
    \label{fig:entire_method}
\end{figure*}

As shown in~\cref{fig:entire_method}, our approach includes three key processes: (1) Memory Bank Construction (top-left): constructing memory banks at three progressive granularities from training captions (\cref{subsec:memory_bank}); (2) Training Process (top-right): retrieving prompts from memory banks using perturbed text embeddings (\cref{subsec:training}); (3) Inference Process (bottom): generating diverse, visually-relevant prompts via category-based retrieval with top-$p$ refinement during inference (\cref{subsec:inference}).

\subsection{Multi-Granularity Memory Bank Construction}\label{subsec:memory_bank}

Our method constructs three distinct memory banks to capture progressive multi-granularity textual semantics, which are used to obtain prompts during the caption generation process: noun phrases, scene graphs incorporating noun phrases, and entire sentences. 
% These memory banks are designed to ensure comprehensive and diverse textual representations of the visual content.

%\paragraph{Noun Phrase Memory Bank (\(\mathcal{M}_{\text{NP}}\))}
\noindent{\textbf{Noun Phrase Memory Bank (\(\mathcal{M}_{\text{NP}}\))}} 
% To construct the noun phrase memory bank, we first extract noun phrases from all captions in the training dataset with SpaCy\footnote{\href{https://spacy.io}{https://spacy.io}}. 
Let \( \mathcal{S} \) represent the set of all captions in the training split. For each caption \( S \in \mathcal{S} \), 
% where \( c_i \) denotes the \( i \)-th caption in the training set, 
we identify all the noun phrases from \( S \) using SpaCy\footnote{\href{https://spacy.io}{https://spacy.io}}, forming a set \( \mathcal{P}(S) \).
% of noun phrases for that specific caption. 
We then aggregate all noun phrases from all captions into a single set. The complete set of noun phrases extracted from the training corpus is denoted as
\begin{equation}
 \mathcal{P} = \bigcup_{S \in \mathcal{S}} \mathcal{P}(S).   
\end{equation}

% where 
% \( \mathcal{P}(c_i) \) represents the set of noun phrases extracted from the \( i \)-th caption \( c_i \), and 
% \( |\mathcal{C}| \) denotes the number of captions in the training set. 
We then rank the noun phrases in \( \mathcal{P} \) based on their frequency of occurrence and retain the top-\( N_p \) most frequent noun phrases to construct the noun phrase memory bank:
\begin{equation}
\mathcal{M}_{\text{NP}} = \{ p_1, p_2, \cdots, p_{N_p} \},
\end{equation}

\noindent where \( p_i \) denotes the \( i \)-th most frequent noun phrase.
% , and \( N_1 \) is the total number of noun phrases retained in the memory bank. 
These noun phrases provide fundamental object-level semantics, serving as the basic building blocks of textual prompts.

%\paragraph{Scene Graph Memory Bank (\(\mathcal{M}_{\text{SG}}\))}
\noindent{\textbf{Scene Graph Memory Bank (\(\mathcal{M}_{\text{SG}}\))}}
% Scene graphs are crucial for capturing the relationships between entities within a video. To build the memory bank, we first utilize an off-the-shelf textual parser~\cite{DBLP:conf/acl/FACTUAL} to extract basic scene graphs from each caption.
% Initially, the results include triples of the subject-predicate-object form, such as \(\langle \text{boy}, \text{play}, \text{basketball} \rangle\). object
% We enhance the scene graph by transforming it from being based solely on nouns to incorporating noun phrases wherever possible. For the objects at the beginning and end of the initial scene graph, if attribute information exists in the caption, we include it in. 
Scene graphs are crucial for capturing the relationships between entities within a video. To build the memory bank, we first utilize an off-the-shelf textual parser~\cite{DBLP:conf/acl/FACTUAL} to extract basic scene graphs from each caption. Initially, the results include triples of the $\langle \textit{subject}, \textit{predicate}, \textit{object} \rangle$ form, such as $\langle \textit{boy}, \textit{play}, \textit{basketball} \rangle$ for the sentence ``A young boy is playing basketball''. We enhance the scene graph by transforming it from being based solely on nouns to incorporating noun phrases wherever possible. For the objects at the beginning and end of the initial scene graph, if attribute information exists in the caption, we include it in. In the example above, we identify ``young boy'' as a noun phrase and transform the initial scene graph into $\langle \textit{young boy}, \textit{play}, \textit{basketball} \rangle$. 
% The entire algorithmic flow is shown in~\cref{alg:sg_memory_v2}.

For the \(i\)-th basic scene graph \(g_i = \langle \textit{sub}_i, \textit{pred}_i, \textit{obj}_i \rangle\) extracted from the caption \(S\), we first identify all noun phrases in \( \mathcal{P}(S) \) that contain \(\textit{sub}_i\),
% as a substring, 
and combine them with \(\textit{sub}_i\) itself to form the set \(\mathcal{A}_i\). Similarly, we form the set \(\mathcal{B}_i\) for \(\textit{obj}_i\). These sets are then used to create the enhanced scene graph set \( \mathcal{X}_i \), where each enhanced graph is a triple of the form \( \langle a, pred_i, b \rangle \), with \( a \in \mathcal{A}_i \) and \( b \in \mathcal{B}_i \).
Next, with the embeddings of \(S\) and all of the enhanced scene graphs in \(\mathcal{X}_i\), denoted as \(\mathbf{E}_S\) and \(\mathbf{E}_{\mathcal{X}_i}\) respectively, all produced by BGE~\cite{BGE} (we encode a scene graph by encoding the string formed by concatenating the subject, predicate, and object with a single space), we calculate the cosine similarity between them. The enhanced scene graph \( x_{\text{best}} \) with the highest cosine similarity to $S$ is selected as the final improved representation of the original scene graph \( g_i \).
% for each enhanced scene graph \( x_i^j \in \mathcal{X}_i \), 
% \[
% x_{\text{best}} = \arg\max_{x \in \mathcal{X}_i} \cos(\mathbf{E}_S, \mathbf{E}_{\mathcal{X}_i}[x])
% \]
Finally, we collect the enhanced scene graphs from all captions and select the top-\( N_g \) most frequent to form the scene graph memory bank \( \mathcal{M}_{\text{SG}} \).  This is used to provide richer and more semantically informative prompts for caption generation.
\begin{equation}
\mathcal{M}_{\text{SG}} = \{ x_1, x_2, \dots, x_{N_g} \},
\end{equation}
\noindent where \( x_i \) denotes the \( i \)-th most frequent enhanced scene graph. We also provide a pseudocode description of the enhanced scene graph memory bank construction in the 
% supplementary material.
appendix.
% (\cref{alg:sg_memory_v2}).

\noindent{\textbf{Entire Caption Memory Bank ($\mathcal{M}_{\text{EC}}$)}}
This memory bank provides holistic textual descriptions that help maintain linguistic coherence in the generated captions. It is simply composed of all captions in the training set.
% , which has already been defined as \( \mathcal{S} \).
% This memory bank provides holistic textual descriptions that help maintain linguistic coherence in the generated captions. The memory bank is simply the set of all captions in the training set, which has already been defined as \( \mathcal{S} \).
% \noindent{\textbf{Entire Caption Memory Bank ($\mathcal{M}_{\text{EC}}$)}}
% This memory bank provides holistic textual descriptions that help maintain linguistic coherence in the generated captions.   
% We define this memory bank as:
% \begin{equation}
% \mathcal{M}_{\text{EC}} = \{ c_1, c_2, ..., c_{N_c} \},
% \end{equation}

% \noindent where \( c_i \) is the the global embedding of \( i \)-th sentence in the training set and \( N_c \) is the total number of training captions. 
% For each caption \( S \) in the training set, we encode it using the text encoder of CLIP~\cite{DBLP:conf/icml/CLIP}.

These three memory banks—noun phrases, enhanced scene graphs, and entire sentences—serve as progressively richer textual representations of the visual content and are significant for guiding the caption generation process.

\subsection{Training Procedure}\label{subsec:training}

% Our training objective is to learn a text decoder capable of generating captions conditioned on multi-granularity prompts, while maintaining robustness to noise in cross-modal retrieval. 
% The text decoder is based on a standard Transformer architecture with text decoder, where each Transformer layer contains self-attention, cross-attention, and feed-forward network (FFN) components. The training process consists of the following four steps. 

Our training objective is to learn a text decoder that generates captions conditioned on multi-granularity prompts, while maintaining robustness to noise during cross-modal retrieval. The text decoder is constructed as a stack of Transformer-BASE~\cite{vanilla_transformer} decoder blocks, each comprising masked self-attention, cross-attention, and feed-forward network (FFN) components. The training process consists of the following four steps.

% we implement MLM with a stack of Transformer decoder blocks, each of which comprises a masked self-attention layer, a crossattention layer, and a feed-forward layer.
% The training process involves three key stages: \textit{(1) Noisy Embedding Augmentation}, \textit{(2) Multi-Scale Prompt Retrieval}, and \textit{(3) Teacher-Forced Decoding}.

% The model is based on a standard Transformer architecture with text decoder, where each Transformer layer contains self-attention, cross-attention, and feed-forward network (FFN) components. During training, we follow the strategy used in MultiCapCLIP and introduce noise to enhance the model's generalization ability, ensuring better transfer from training to inference.
%\paragraph{Noisy Embedding Augmentation}
\noindent{\textbf{Step 1: Embedding Augmentation}}
For each caption \( S_o \), we first retrieve the most similar \( M \) captions from the training set, based on their cosine similarity to the CLIP sentence embedding \( \mathbf{e}(S_o) \). Among these \( M \) captions, we randomly select one, denoted as \( S_r \). We use \( \mathbf{e}(S_r) \) and add Gaussian noise \(\epsilon \sim \mathcal{N}(0, \lambda^2)\) to obtain a perturbed embedding \(\tilde{\mathbf{e}}\):
\begin{equation}
\tilde{\mathbf{e}} = \mathbf{e}(S_r) + \epsilon.
\end{equation}

%\paragraph{Memory Bank Retrieval}\\\\TMK from here
\noindent{\textbf{Step 2: Memory Bank Retrieval}}
% The perturbed embedding \(\tilde{\mathbf{e}}\) is then used to retrieve the top-\(K_p\) noun phrases and the top-\(K_g\) scene graphs from their respective memory banks whose embeddings, encoded by CLIP, have the highest cosine similarity with \(\tilde{\mathbf{e}}\).
The perturbed embedding \(\tilde{\mathbf{e}}\) is used to retrieve the top-\(K_p\) noun phrases and top-\(K_g\) scene graphs from \(\mathcal{M}_{\text{NP}}\) and \(\mathcal{M}_{\text{SG}}\), respectively, based on the cosine similarity of CLIP embeddings. The representations of these retrieved elements are denoted as \(\mathbf{e}_{\text{NP}}\) and \(\mathbf{e}_{\text{SG}}\).

% We retrieve the top \( K_1 \) noun phrases and top \( K_2 \) scene graphs :

% \[
% \mathcal{P}_{\text{retrieved}} = \text{Top-K1}(\text{cosine}(\mathbf{e}, \mathcal{M}_{\text{NP}}))
% \]
% \[
% \mathcal{G}_{\text{retrieved}} = \text{Top-K2}(\text{cosine}(\mathbf{e}, \mathcal{M}_{\text{SG}}))
% \]

% where \( \mathcal{M}_{\text{NP}} \) and \( \mathcal{M}_{\text{SG}} \) represent the noun phrase memory bank and the scene graph memory bank, respectively. The retrieved noun phrases and scene graphs, along with the original sentence embedding, are processed by their respective FFN layers.
%\paragraph{Prompt Construction}
\noindent{\textbf{Step 3: Prompt Construction}}
\(\mathbf{e}_{\text{NP}}\), \(\mathbf{e}_{\text{SG}}\), and \(\tilde{\mathbf{e}}\) are passed through individual FFNs and then concatenated as the prefix prompt for the text decoder. Specifically, the final prompt is given by:
\begin{equation}
\mathbf{P} = \texttt{Concat}(\mathbf{e}_{\text{NP}}^{\prime}, \mathbf{e}_{\text{SG}}^{\prime}, \mathbf{e}_{\text{EC}}),
\end{equation}

\noindent where \( \mathbf{e}_{\text{NP}}^{\prime} \), \( \mathbf{e}_{\text{SG}}^{\prime} \), and \( \mathbf{e}_{\text{EC}} \) represent the transformed embeddings of the noun phrases, scene graphs, and the entire sentence, respectively.

%\paragraph{Teacher-Forced Decoding}
\noindent{\textbf{Step 4: Teacher-Forced Decoding}}
The model is trained using teacher forcing, 
% where the true target sentence  is used as the input during training. 
and the goal is to minimize the cross-entropy loss:
% between the predicted output and the true target sentence \( S_o \):
\begin{equation}
\mathcal{L} = - \sum_{t=1}^{T} \log p(y_t | y_{<t}, \mathbf{P}; \theta),
\end{equation}

\noindent where \( y_t \) is the target word of sentence \( S_o \) at time step \( t \), 
% \( \mathbf{y}_{t-1} \) is the previous word in the sequence, \( \mathbf{P} \) is the concatenated prefix prompt, 
and \( \theta \) represents the model parameters. 
% The model is optimized to predict the next word in the sequence by minimizing the loss.

\subsection{Inference with Category-Aware Retrieval}\label{subsec:inference}

During inference, we utilize a specialized strategy for prompt generation, combining category-aware retrieval with top-\textit{p} post-processing, to ensure diverse and relevant textual prompts for accurate and expressive video captions.

% During inference, we employ a specialized strategy for prompt generation, leveraging category-aware retrieval with top-\textit{p} post-processing. This approach ensures that the textual prompts are diverse and highly relevant to the visual input, enabling the model to generate accurate and expressive video captions.

\subsubsection{Noun Phrase Prompt Generation}
\label{Noun Phrase Prompt Generation}

The inference begins with generating prompts from \( \mathcal{M}_{\text{NP}} \), involving noun phrase classification, relevant candidates retrieval based on statistical priors, and refinement with a top-\textit{p} mechanism. Specifically, the retrieval step employs statistical priors tailored to the in-domain and cross-domain settings, respectively, and is discussed separately below.

\noindent{\textbf{Classification with GPT-4}}
For the noun phrase memory bank \( \mathcal{M}_{\text{NP}} \), we apply GPT-4~\cite{gpt4} to automatically classify the noun phrases into different categories. The classification is completely unsupervised, allowing GPT-4 to determine the optimal categories for the noun phrases. Once the classification is completed, all noun phrases are assigned to one of the predefined categories, such as singular people, object, place, etc. Distribution details are provided in the 
% supplementary material.
appendix.
% (\cref{Classification_Result_of_NP_Memory_Bank}).

%\paragraph{In-Domain Retrieval with Statistical Priors}
\noindent{\textbf{In-Domain Retrieval with Statistical Priors}}
To adaptively determine hyperparameters (\eg, the number of most relevant elements to select per-category) and obviate manual configuration during the categorized retrieval process, we initially compute in-domain statistical priors. For a video from the training video set \( \mathcal{V} \), unique noun phrases from its corresponding captions form a set \( \mathcal{P}_V \). For each category \( \kappa \) with noun phrases \( \mathcal{P}_{\kappa} \subseteq \mathcal{M}_{\text{NP}} \), we compute two statistics: a probability of occurrence, \( p_{\kappa} = \frac{N_{\kappa}}{|\mathcal{V}|} \), where \( N_{\kappa} \) is the number of videos containing at least one noun phrase in \( \mathcal{P}_{\kappa} \), and \( |\mathcal{V}| \) is the total number of training set videos; and an average frequency, \( \mu_{\kappa} = \frac{N_{\kappa}^{\mathcal{P}}}{N_{\kappa}} \), where \( N_{\kappa}^{\mathcal{P}} \) is the total count of noun phrases parsed from the training corpus that overlap with \( \mathcal{P}_{\kappa} \). This process is formalized in~\cref{Category-based Statistics Computation}.

% For each category \( \kappa \) with noun phrases \( \mathcal{P}_{\kappa} \subseteq \mathcal{M}_{\text{NP}} \), we compute two statistics:

% \begin{itemize}
%     \item Probability of occurrence: \( p_{\kappa} = \frac{N_{\kappa}}{|\mathcal{V}|} \), where \( N_{\kappa} \) is the number of videos containing at least one noun phrase in \( \mathcal{P}_{\kappa} \), and \( |\mathcal{V}| \) is the total number of training set videos.
%     \item Average frequency: \( \mu_{\kappa} = \frac{N_{\kappa}^{\mathcal{P}}}{N_{\kappa}} \), where \( N_{\kappa}^{\mathcal{P}} \) is the total count of noun phrases parsed from training corpus that overlap with \( \mathcal{P}_{\kappa} \).
% \end{itemize}

Next, for a test video \( V = \{f_t\}_{t=1}^T \), we compute frame-level cosine similarity between its CLIP visual features \(\phi(f_t) \) and text embeddings \( \mathbf{e}(n) \) of noun phrase \( n \in \mathcal{P}_{\kappa} \). The video-phrase similarity $s_n$ is derived by averaging frame-level scores. We retrieve the top-\( \texttt{round}(\mu_{\kappa}) \) noun phrases from \( \mathcal{P}_{\kappa} \) based on \( s_n \), and retain all of them with probability \( p_{\kappa} \). The retained noun phrases across all categories are aggregated into a set \( Y \).
% \noindent Next, we retrieve noun phrases from each category based on above statistics. Given a test video \( V = \{f_t\}_{t=1}^T \), where \( f_t \) denotes the \( t \)-th frame, we extract visual features for each frame using CLIP (ViT/B-16). For each category \( \kappa \), we:
% \begin{itemize}
%     \item Compute the frame-level cosine similarity \( s_{t,n} = \cos(\phi(f_t), \mathbf{e}(n))\) for each frame \( f_t \) and each noun phrase \( n \in \mathcal{P}_{\kappa} \), where \( \phi(f_t) \) is the CLIP visual embedding of frame \( f_t \), and \( \mathbf{e}(n) \) is the CLIP text embedding of \( n \). Following MultiCapCLIP, the video-level similarity is then obtained by averaging across frames: \( s_n = \frac{1}{T} \sum_{t=1}^T s_{t,n} \).
%     \item Retrieve the top-\( \texttt{round}(\mu_{\kappa}) \) noun phrases with the highest \( s_n \), and retain all of them with probability \( p_{\kappa} \).
%     % reflecting their prevalence in the training data.
%     % \item Retain these noun phrases with probability \( p_{\kappa} \), reflecting their prevalence in the training data.
% \end{itemize}
% The retained noun phrases across all categories are aggregated into a set \( Y \).

%\paragraph{Cross-Domain Retrieval with Statistical Priors}\label{Cross-Domain Retrieval with Text-derived Priors}
\noindent{\textbf{Cross-Domain Retrieval with Statistical Priors}}
In the cross-domain scenario, visual features of the target domain are leveraged to retrieve textual units from memory banks constructed from the source domain. These retrieved prompts are subsequently fed into the text decoder pre-trained on the source domain to generate captions.
We compute category statistics using only the source domain training captions \( \mathcal{S} \). For each category \( \kappa \), we compute the total count of noun phrases parsed from the training corpus that belong to the category's noun phrase set \( \mathcal{P}_{\kappa} \), denoted as \( N_{\kappa} \).
We designate the category with the minimum count \( N_{\kappa} \) as the base category, with its corresponding count denoted as \( b \). 
Then, for each category \( \kappa \), we retrieve  \( r_{\kappa} = \texttt{round}(N_{\kappa} / b \cdot B) \) noun phrases, where \( B \) is a pre-defined base retrieval number. We adopt the same cross-modal similarity computation method as in the in-domain setting and aggregate the top-\(r_{\kappa}\) most relevant elements from each category into a set \( Y \).

%\paragraph{Top-\textit{p} Refinement}
\noindent{\textbf{Top-\textit{p} Refinement}}
% To balance relevance and diversity, we refine \( Y \) using a top-\textit{p} strategy. The similarity \( s_n \) is normalized into a probability distribution $\hat{s}_n$ and we select the smallest set of noun phrase $Y'$ whose combined probability exceeds a threshold p (e.g.,$\tau$), allowing for a more context-sensitive choice of noun phrase.
To balance relevance and diversity, we refine \(Y\) using a top-\(p\) strategy~\cite{Top_p}: we normalize the similarities between the video and all noun phrases in \(Y\) into a probability distribution \(\{\hat{s}_n | n \in Y \}\), sort \(Y\) in descending order of \(\hat{s}_n\), and select the smallest subset \(Y' \subseteq Y\) whose cumulative probability \(\sum_{n \in Y'} \hat{s}_n\) reaches a predefined threshold \(\tau\).
%\begin{itemize}
%    \item Reuse \( s_n \) computed in the last step for all \( n \in Y \)
    % , where \( s_n = \frac{1}{T} \sum_{t=1}^T s_{t,n} \) and \( s_{t,n} = \cos(\phi(f_t), \mathbf{e}(n)) \).
%    \item Normalize these scores into a probability distribution
    % : \( \hat{s}_n = \frac{s_n}{\sum_{n' \in Y} s_{n'}} \), where the denominator is the sum of all similarity scores in \( Y \).
%    \item Sort \( Y \) by \( \hat{s}_n \) in descending order and select the smallest subset \( Y' \subseteq Y \) such that \( \sum_{n \in Y'} \hat{s}_n \geq \tau \), where \( \tau \) is a predefined probability threshold.
%\end{itemize}
We encode \( Y' \) using CLIP text encoder to obtain the noun phrase prompt \( \mathbf{e}(Y') \) of test video $V$.

\begin{algorithm}[t]
\small
\SetAlgoLined
\KwIn{$\mathcal{V}$: Training videos; $\mathcal{P}_{\kappa}$: Noun phrases of category $\kappa$}
\KwOut{$\mu_{\kappa}$, $p_{\kappa}$: Average number of noun phrases per video and category probability}
\( N_{\kappa} \gets 0 \), \( N_{\kappa}^{\mathcal{P}} \gets 0 \)\;
\For{$V \in \mathcal{V}$}{
    Parse captions of $V$ to extract noun phrases $\mathcal{P}_V$\;
    \( \mathcal{P}_V \gets \text{Remove duplicates from } \mathcal{P}_V \)\;
    \If{\( \mathcal{P}_V \cap \mathcal{P}_{\kappa} \neq \emptyset \)}{
        \( N_{\kappa} \gets N_{\kappa} + 1 \)\;
        \( N_{\kappa}^{\mathcal{P}} \gets N_{\kappa}^{\mathcal{P}} + |\mathcal{P}_V \cap \mathcal{P}_{\kappa}| \)\;
    }
}
\( p_{\kappa} \gets \frac{N_{\kappa}}{|\mathcal{V}|} \), \( \mu_{\kappa} \gets \frac{N_{\kappa}^{\mathcal{P}}}{N_{\kappa}} \)\;
\Return \( p_{\kappa}, \mu_{\kappa} \)
\caption{Category-based Statistics Computation}
\label{Category-based Statistics Computation}
\end{algorithm}

% \paragraph{Cross-Domain Retrieval with Text-derived Priors}
% \label{Cross-Domain Retrieval with Text-derived Priors}
% In the cross-domain setting, we compute category statistics using only the source domain training captions \( \mathcal{S} \). For each category \( \kappa \), we compute the total count of noun phrases parsed from the training corpus that belong to \( \kappa \), denoted as \( N_{\kappa} \).

% We then determine a base count \( \text{base} \) by selecting the minimum \( N_{\kappa} \) across all categories. For each category \( \kappa \), we compute a scaling factor \( r_{\kappa} = \text{round}(N_{\kappa} / \text{base}) \).

% Then, for each category \( \kappa \), we retrieve the top \( r_{\kappa} \) noun phrases with the highest cosine similarity to the video's visual features.

\subsubsection{Scene Graph Prompt Generation}
For the scene graph prompt generation, we follow a pipeline similar to the previous section: Scene graphs in \( \mathcal{M}_{\text{SG}} \) are classified by pairing the categories of their subject and object noun phrases (\eg, ``People\_\textit{pred}\_Object'' in~\cref{fig:entire_method}). Noun phrases not in \( \mathcal{M}_{\text{NP}} \) are assigned to the category of their nearest neighbor in \( \mathcal{M}_{\text{NP}} \) based on BGE embedding similarity. Then, we computed the statistical priors analogously to the noun phrase case and aggregate the retrieved items into a set \( Z \). Finally, top-\textit{p} filtering is applied to \( Z \), and the filtered result is encoded with CLIP’s text encoder to produce the scene graph prompt \( \mathbf{e}(Z') \).

\subsubsection{Entire Caption Prompt Generation}
% For the entire caption prompt, we adopt a DeCap-inspired approach:

% \begin{itemize}
%     \item Compute the CLIP similarity \( s_c = \text{CLIP}(\phi(V_i), e_c) \) between \( \phi(V_i) \) and the embeddings \( e_c \) of all captions \( c \in \mathcal{M}_{\text{EC}} \).
%     \item Apply softmax to obtain weights: \( w_c = \frac{\exp(s_c / \tau)}{\sum_{c' \in \mathcal{M}_{\text{C}}} \exp(s_{c'} / \tau)} \).
%     \item Generate a single prompt token: \( P_{\text{C}} = \sum_{c \in \mathcal{M}_{\text{C}}} w_c \cdot e_c \).
% \end{itemize}

For the entire caption prompt, we  compute the similarity between each video and each global caption embedding in \( \mathcal{M}_{\text{EC}} \) as in DeCap~\cite{DBLP:conf/iclr/DeCap},
% based on CLIP embeddings, 
then apply softmax to obtain weights, and  finally generate a single prompt token \( \mathbf{e}_c \) by weighted summation.

\subsubsection{Integrating Prompts and Generating Captions}
\( \mathbf{e}(Y') \), \( \mathbf{e}(Z') \), and \( \mathbf{e}_c \) are processed by their respective FFNs and concatenated into a sequence, which is fed into our text decoder to generate the final caption for \( V \).

\section{Experiments}

We begin by introducing the datasets, evaluation metrics, and implementation details in~\cref{Experimental_Setups}.~\cref{theAnalysis} presents comprehensive comparisons with existing methods on in-domain zero-shot video captioning, wherein the model undergoes both training and evaluation on the same dataset.~\cref{Cross-domain} further evaluates our method on cross-domain scenario, using a source domain corpus for training and a target domain dataset for evaluation.~\cref{ablation} conducts ablation studies to validate the effectiveness of our core design and the scalability of our model. Finally,~\cref{Qualitative} provides qualitative visualizations that intuitively demonstrate the superior performance of our model.

\subsection{Experimental Setups}\label{Experimental_Setups}

\noindent{\textbf{Datasets}} We evaluate our method on three video-text datasets: MSR-VTT~\cite{DBLP:conf/cvpr/MSR-VTT}, MSVD~\cite{DBLP:conf/acl/MSVD}, and VATEX~\cite{VATEX}. MSR-VTT includes 10,000 web videos, split into 6,513 training, 497 validation, and 2,990 testing videos. MSVD contains 1,970 YouTube clips, divided into 1,200 training, 100 validation, and 670 testing videos. VATEX comprises over 30,000 videos, and we use 25,006 clips for training, and 2,893 and 5,792 video clips for validation and testing respectively following MultiCapCLIP~\cite{DBLP:conf/acl/MultiCapCLIP}. Experiments on VATEX concentrate exclusively on the English corpus.

%\paragraph{Evaluation Metrics}
\noindent{\textbf{Evaluation Metrics}}
Following the common practice, we evaluate the caption quality with four metrics, including BLEU@4 (B@4)~\cite{DBLP:conf/acl/BLEU4}, METEOR (M)~\cite{DBLP:conf/acl/METEOR}, ROUGE-L (R)~\cite{DBLP:conf/acl/ROUGE-L} and CIDEr (C)~\cite{DBLP:conf/cvpr/CIDEr}. Among them, the \textbf{CIDEr} is specifically designed to evaluate captioning systems and better captures human judgment of consensus better than the others. We also report Self-BLEU~\cite{self_BLEU}, a measure of text diversity in our ablation study on retrieval strategies.

%\paragraph{Implementation Details}
\noindent{\textbf{Implementation Details}}
In~\cref{tab:hyper}, we present key hyperparameters in our method.
For feature extraction, we employ the frozen pre-trained CLIP (ViT/B-16).
% vision-language model to encode visual and textual inputs.
The text decoder is an 6-layer Transformer trained from scratch. 
% More details are provided in supplementary material. 
More details are provided in appendix. 

\begin{table}[tbp]
  \centering
  \resizebox{\linewidth}{!}{
    \begin{tabular}{cccccccc}
    \toprule
      & $ N_p $ & $ N_g $ & $ N_c $ & $ K_p $ & $ K_g $ & $ M $ & $\lambda^2$ \\
    \midrule
    MSVD     & 1000  & 37711  & 48774  & 13    & 16    & 5 & 0.01\\
    MSR-VTT  & 1000  & 100000 & 130260 & 14    & 19    & 5 & 0.01\\
    VATEX    & 3000  & 400000 & 250060 & 10    & 13    & 5 & 0.01\\
    \bottomrule
    \end{tabular}}
    \caption{Hyperparameters used in our in-domain experiments. \( N_c \) represents the number of captions in each dataset’s training set. For MSVD, 37711 equals the total number of enhanced scene graphs derived from all training captions.}
  \label{tab:hyper}%
\end{table}%

\subsection{In-domain Captioning}\label{theAnalysis}
As shown in \cref{tab:indomainMSRVTT,tab:indomainVATEX}, our method establishes new state-of-the-art zero-shot performance across all benchmarks. Three key performance patterns emerge:

\noindent{\textbf{Vertical Dominance}}
Our method significantly outperforms existing text-only training methods, achieving CIDEr scores of 39.3\% and 92.9\% on MSR-VTT and MSVD, respectively, exceeding MultiCapCLIP by 5.7\% and 16.2\%. Superior performance is also observed on VATEX across all metrics. These results, coupled with substantial gains over training-free methods, validate the efficacy of our approach.
% The proposed method exhibits superior performance compared to existing text-only training approaches. Specifically, CIDEr scores of 39.3% and 92.9% are achieved on MSR-VTT and MSVD, representing improvements of 5.7% and 16.2% over MultiCapCLIP, respectively. Furthermore, our method outperforms both DeCap and MultiCapCLIP across all metrics on the VATEX dataset. These findings, along with considerable improvements over training-free baselines, underscore the effectiveness of the proposed approach.

\noindent{\textbf{Horizontal Consistency}}
The disparity in CIDEr scores between MSVD (92.9\%) and VATEX (41.4\%) reflects the positive correlation between absolute performance and dataset complexity. VATEX, with its longer videos and denser temporal relations, presents a greater challenge than the shorter, simpler clips in MSVD. 
% This suggests that while our method achieves strong performance across datasets, there remains room for improvement in handling complex temporal dynamics.
% \noindent{\textbf{Horizontal Consistency}}
% Performance gaps between datasets (CIDEr: 41.4 on VATEX vs. 92.9 on MSVD) correlate with video complexity. VATEX's dense temporal relations and longer videos pose greater challenges than MSVD's shorter, simpler clips. This suggests that while our method excels across datasets, there's still room for improvement in handling complex temporal dynamics.

% \noindent{\textbf{Supervised Proximity}}
% On MSVD, our CIDEr score of 92.9 not only surpasses many supervised methods (e.g., SAAT, STR, POS-CG) but also approaches the performance of VPT (94.7\%), a strong supervised baseline. Although a performance gap remains on MSR-VTT and VATEX compared to top supervised methods, our results underscore the potential of our approach to narrow the gap with traditional supervised techniques, without relying on expensive video-text pair annotations.

\noindent{\textbf{Supervised Proximity}}
On MSVD, our CIDEr score of 92.9\% has already exceeded supervised methods like SAAT~\cite{DBLP:conf/cvpr/SAAT}, STR~\cite{STR}, and POS-CG~\cite{DBLP:conf/iccv/POS-CG}, and is approaching VPT's~\cite{VPT} 94.7\%. Although a performance gap remains on MSR-VTT and VATEX compared to top supervised methods, our results highlight the strong potential of our approach in narrowing the gap with traditional supervised techniques.

\begin{table*}[htbp]
  \centering
  \resizebox{\textwidth}{!}{
    \begin{tabular}{ccccccccccccc}
    \toprule
    \multirow{2}[4]{*}{Settings} & \multirow{2}[4]{*}{Method} & \multirow{2}[4]{*}{Pre-trained Model} & \multicolumn{2}{c}{Training Data} & \multicolumn{4}{c}{MSR-VTT}   & \multicolumn{4}{c}{MSVD} \\
\cmidrule(lr){4-5}\cmidrule(lr){6-9}\cmidrule(lr){10-13}          &       &       & Video & Text  & B@4   & M     & R     & C     & B@4   & M     & R     & C \\
    \midrule
    \multirow{7}[2]{*}{Supervised} & SGN~\cite{DBLP:conf/aaai/SGN}   & ResNet-101 + C3D & \checkmark     & \checkmark     & 40.8  & 28.3  & 60.8  & 49.5  & 52.8  & 35.5  & 72.9  & 94.3 \\
          & POS-CG~\cite{DBLP:conf/iccv/POS-CG} & InceptionResNetV2 & \checkmark     & \checkmark     & 42.0    & 28.2  & 61.6  & 48.7  & 52.5  & 34.1  & 71.3  & 88.7 \\
          & SAAT~\cite{DBLP:conf/cvpr/SAAT}  & InceptionResNetV2 + C3D & \checkmark     & \checkmark     & 39.9  & 27.7  & 61.2  & 51.0    & 46.5  & 33.5  & 69.4  & 81.0 \\
          & HRNAT~\cite{DBLP:journals/tip/HRNAT} & InceptionResNetV2 + I3D & \checkmark     & \checkmark     & 42.1  & 28.0    & 61.6  & 48.2  & 55.7  & 36.8  & 74.1  & 98.1 \\
          & STR~\cite{STR} & InceptionResNetV2 + I3D & \checkmark     & \checkmark     & - & 25.8 & 54.8 & 47.6  &  - & 34.2& 68.6& 86.5 \\
          & VPT~\cite{VPT}   & CLIP (ViT/B-16) & \checkmark     & \checkmark     & 41.2  & 27.9  & 61.5  & 50.3  & 54.6  & 36.0    & 73.1  & 94.7 \\
          % & Prompting~\cite{Prompting} & CLIP (ViT/B-16) & \checkmark     & \checkmark     & 42.0    & 28.8  & 62.3  & 52.2  & 56.4  & 38.6  & 75.4  & 99.8 \\
          & CoCap~\cite{CoCap} & CLIP (ViT/B-16) & \checkmark     & \checkmark     & 43.1  & 29.8  & 62.7  & 56.2  & 55.9  & 39.9  & 76.8  & 113.0 \\
    \midrule
    \multirow{6}[2]{*}{Zero-shot} & ZeroCap~\cite{DBLP:conf/cvpr/ZeroCap} & CLIP (ViT/B-32) + GPT & $\times$     & $\times$     & 2.3   & 12.9  & 30.4  & 5.8   & 2.9   & 16.3  & 35.4  & 9.6 \\
          & ZeroCap-Video~\cite{DBLP:conf/bmvc/ZeroCap_video} & CLIP (ViT/B-32) + GPT & $\times$     & $\times$     & 3.0     & 14.6  & 27.7  & 11.3  & 3.0     & 17.8  & 31.4  & 17.4 \\
          & MAGIC~\cite{DBLP:journals/corr/MAGIC} & CLIP (ViT/B-32) + GPT & $\times$     & \checkmark     & 5.5   & 13.3  & 35.4  & 7.4   & 6.6   & 16.1  & 40.1  & 14.0 \\
          & DeCap\ddag~\cite{DBLP:conf/iclr/DeCap} & CLIP (ViT/B-16) & $\times$     & \checkmark     & 26.6  & 23.5  & 53.2  & 29.7  & 35.2  & 29.0    & 65.2  & 41.3 \\
          & MultiCapCLIP\dag~\cite{DBLP:conf/acl/MultiCapCLIP} & CLIP (ViT/B-16) & $\times$     & \checkmark     & 22.0    & 24.4  & 50.2  & 33.6  & 40.2  & 34.2  & 68.6  & 76.7 \\
          & \textbf{Ours}  & CLIP (ViT/B-16) & $\times$     & \checkmark     & \textbf{31.4}  & \textbf{26.5}  & \textbf{55.1}  & \textbf{39.3}  & \textbf{45.7}  & \textbf{35.9}  & \textbf{71.5}  & \textbf{92.9} \\
    \bottomrule
    \end{tabular}%
    }
 \caption{
 In-domain captioning results on the MSR-VTT and MSVD test sets. \ddag indicates the reproduced results on both datasets using CLIP (ViT/B-16) for fair comparison. \dag ~denotes that the results on MSVD are from our implementation. 
 % The best scores are highlighted in \textbf{bold}.
  }
  \label{tab:indomainMSRVTT}%
\end{table*}%

% In-Domain Video Captioning Performance on MSR-VTT and MSVD Test Sets. \ddag: DeCap reproduction with CLIP (ViT/B-16) for fair comparison; \dag: MultiCapCLIP implementation on MSVD. Best scores in \textbf{bold}.

\begin{table}[htbp]
  \centering
  \resizebox{\linewidth}{!}{
    \begin{tabular}{cccccc}
    \toprule
    \multirow{2}[4]{*}{Settings} & \multirow{2}[4]{*}{Method} & \multicolumn{4}{c}{VATEX} \\
\cmidrule{3-6}          &       & B@4   & M     & R     & C \\
    \midrule
    \multirow{3}[2]{*}{Supervised} & VATEX~\cite{VATEX} & 28.4  & 21.7  & 47.0    & 45.1 \\
          % & STR~\cite{STR}   & -     & 20.1  & 43.6  & 45.0 \\
          & HRNAT~\cite{DBLP:journals/tip/HRNAT} & 32.5  & 22.3  & 49.0    & 50.7 \\
          & CoCap~\cite{CoCap} & 31.4  & 23.2  & 49.4  & 52.7 \\
    \midrule
    \multirow{3}[2]{*}{Zero-shot} & DeCap\ddag~\cite{DBLP:conf/iclr/DeCap} & 19.2  & 19.3  & 42.8  & 27.5 \\
          & MultiCapCLIP\dag~\cite{DBLP:conf/acl/MultiCapCLIP} & 21.7  & 20.1  & 43.3  & 38.0 \\
          & \textbf{Ours}  & \textbf{23.8}  & \textbf{21.0}    & \textbf{44.5}  & \textbf{41.4} \\
    \bottomrule
    \end{tabular}%
    }
    \caption{In-domain captioning results on the VATEX test set. \ddag~ indicates the reproduced results using CLIP (ViT/B-16). \dag ~marks the MultiCapCLIP implementation with English annotations. 
    % The best scores are highlighted in \textbf{bold}.
    }
  \label{tab:indomainVATEX}%
\end{table}%

\begin{table}[tbp]
  \centering
  \resizebox{\linewidth}{!}{
    \begin{tabular}{ccccccccc}
    \toprule
    \multirow{2}[4]{*}{Method} & \multicolumn{4}{c}{MSR-VTT $\Rightarrow$ MSVD} & \multicolumn{4}{c}{MSVD $\Rightarrow$ MSR-VTT} \\
\cmidrule(lr){2-5}  \cmidrule(lr){6-9}        & B@4   & M     & R     & C     & B@4   & M     & R     & C \\
    \midrule
    DeCap\ddag & 23.1  & 25.4  & 56.9  & 28.2  & 16.4  & 18.6  & 50.3  & 8.8 \\
    MultiCapCLIP\ddag &  24.8     & 28.9      & 57.7      &  39.9     & 20.4      & 22.2      & 50.9      & 22.4 \\
    \textbf{Ours}  &  \textbf{28.9}     & \textbf{30.4}      & \textbf{60.6}      &   \textbf{51.7}    &  \textbf{25.0}     & \textbf{23.2}      & \textbf{54.7}     & \textbf{28.1}  \\
    \bottomrule
    \end{tabular}}
    \caption{Performance on cross-domain captioning. \ddag\ denotes our reproduction with CLIP (ViT/B-16).}
  \label{tab:cross-domain}%
\end{table}%

\subsection{Cross-domain Captioning}\label{Cross-domain}

To evaluate the generalization ability of our method, we conduct cross-domain experiments on MSR-VTT and MSVD datasets. The results are presented in~\cref{tab:cross-domain}.
Our method consistently outperforms both DeCap and MultiCapCLIP. In MSR-VTT \(\Rightarrow\) MSVD task, we achieve a CIDEr score of 51.7\% (+11.8\% over MultiCapCLIP, +23.5\% over DeCap) and a B@4 score of 28.9\% (+4.1\% over MultiCapCLIP, +5.8\% over DeCap).  In MSVD \(\Rightarrow\) MSR-VTT task, we obtain a CIDEr score of 28.1\% and a B@4 score of 25.0\%, again demonstrating significant enhancements.
These improvements underscore the superior generalization performance of our approach, attributed to the utilization of multi-granularity textual semantic units and category-aware retrieval with top-\(p\) filtering. This facilitates effective knowledge transfer from source to target domain without video-text pairs.

\subsection{Ablation studies}\label{ablation}

\noindent{\textbf{Multi-granularity Textual Prompts}}
To evaluate the contributions of our proposed progressive multi-granularity prompts, we conduct in-domain ablation studies on MSR-VTT and MSVD, with results in~\cref{tab:ablation_multiGrained}. Starting with \texttt{Ours (w/o Prompt)}, adding noun phrase prompts (\texttt{Ours (NP)}) boosts performance across all metrics—\eg, BLEU@4 increases from 25.9\% to 28.7\% on MSR-VTT and CIDEr from 60.8\% to 79.7\% on MSVD—highlighting their role in providing key entity information. 
Incorporating scene graphs enhanced with noun phrases (\texttt{Ours (NP+SG)}) yields further gains, notably a 9.5\% CIDEr increase on MSVD, capturing relational and action details. The full model (\texttt{Ours (NP+SG+EC)}), integrating entire caption prompts, achieves peak performance—\eg, BLEU@4 of 31.4\% on MSR-VTT and CIDEr of 92.9\% on MSVD—demonstrating that the progressive inclusion of multi-granularity prompts continuously refines caption quality by capturing complementary semantic information at various abstraction levels.

\noindent{\textbf{Category-aware Retrieval with Top-\textit{p}}}
We evaluate the effectiveness of our retrieval strategy for cross-domain captioning using noun phrase prompts, with results in~\cref{tab:ablation_retrieval}. Both \texttt{Direct Top-K} and \texttt{Category-based} methods retrieve the same total number of noun phrases: the former selects the most similar ones from the entire memory bank, while the latter retrieves specific number from each category based on statistical priors. 
% The \texttt{Category-based Retrieval with Top-\textit{p} Filtering} method refines this set using Top-\textit{p} post-processing. 
The \texttt{Category-based} method improves diversity of retrieved noun phrases over \texttt{Direct Top-K}, as evidenced by a lower self-BLEU score, though it slightly reduces CIDEr due to potential noise from irrelevant categories. Adding top-\textit{p} filtering further enhances diversity
% markedly improves caption quality, with CIDEr rising to 23.2 on MSVD $\Rightarrow$ MSR-VTT and 41.4 on MSR-VTT $\Rightarrow$ MSVD. This demonstrates that our category-aware retrieval with Top-\textit{p} filtering effectively balances diversity and relevance, yielding superior captions.
and significantly boosts caption quality across all metrics, with CIDEr improving from 31.0 to 41.4\% on MSR-VTT $\Rightarrow$ MSVD task. These results highlight the superiority of our category-aware retrieval combined with top-\textit{p} filtering, which balances diversity and relevance to produce higher-quality captions.

%\paragraph{Scaling Up Pre-trained Multimodal Models}\label{scaleup_quanti}
\noindent{\textbf{Scaling Up Pre-trained Multimodal Models}}
In~\cref{tab:Pre_trained_VL_Scaling}, we explore how the scale of pre-trained multimodal models affects the performance of zero-shot video captioning on two relatively large datasets, MSR-VTT and VATEX. As model size increases, all metrics improve significantly on both datasets.  For instance, on MSR-VTT, scaling from CLIP (ViT/B-16) to GME-Qwen2VL-7B~\cite{DBLP:journals/corr/GME} increases the CIDEr score from 39.3\% to 48.2\%. The improvement is even more pronounced on VATEX, with CIDEr increasing from 41.4\% to 62.2\% (+20.8\%).
This indicates that our approach, when supported by larger multimodal models, can more more effectively retrieves relevant textual elements, achieving notable gains under a text-only training, visual-only inference paradigm.  The advantage of stronger pre-trained vision-language models is particularly evident in the more complex VATEX dataset, highlighting our method's potential with even more powerful multimodal models in the future. 
% See the supplementary material 
See the appendix 
% (\cref{scaleQuali}) 
for a qualitative comparison of retrieved text units and generated captions across model scales.

% For a detailed qualitative comparison of retrieved text units and generated captions across model scales, please refer to the supplementary material (\cref{scaleQuali}).
% This qualitative observation aligns with the quantitative improvements discussed earlier and underscores the enhanced ability of larger multimodal models to extract and integrate finer-grained semantic information, thus improving caption quality.

%\begin{table}[t]
%\centering
%\resizebox{0.8\linewidth}{!}{
%\begin{tabular}{ccccc}
%\toprule
%Method & B@4 & M & R & C \\
%\midrule
%\multicolumn{5}{c}{MSR-VTT} \\
%\hline
%Ours (w/o Prompt) & 25.9    & 25.8  & 52.6  & 29.8 \\
%Ours (NP) & 28.7 & 26.0 & 54.0 & 33.9 \\
%Ours (NP + SG) & 29.7 & 26.2 & 54.8 & 37.3 \\
%Ours (NP + SG + EC) & \textbf{31.4}  & \textbf{26.5}  & \textbf{55.1}  & \textbf{39.3} \\
%\midrule
%\multicolumn{5}{c}{MSVD} \\
%\hline
%Ours (w/o Prompt) & 33.4  & 31.4  & 64.9  & 60.8 \\
%Ours (NP) & 40.9 & 34.1 & 68.7 & 79.7 \\
%Ours (NP + SG) & 45.0 & 35.6 & 71.1 & 89.2 \\
%Ours (NP + SG + EC) & \textbf{45.7}  & \textbf{35.9}  & \textbf{71.5}  & \textbf{92.9} \\
%\bottomrule
%\end{tabular}}
%\caption{Ablation study on multi-granularity semantic prompts. NP: Noun Phrase, SG: Scene Graph, EC: Entire Caption.}
%\label{tab:ablation_multiGrained}
%\end{table}

\begin{table}[tbp]
  \centering
  \resizebox{\linewidth}{!}{
    \begin{tabular}{ccccccccc}
    \toprule
    \multirow{2}[4]{*}{Method} & \multicolumn{4}{c}{MSR-VTT } & \multicolumn{4}{c}{MSVD } \\
\cmidrule(lr){2-5}  \cmidrule(lr){6-9}        & B@4   & M     & R     & C     & B@4   & M     & R     & C \\
    \midrule
   Ours (w/o prompt)&  25.9    & 25.8  & 52.6  & 29.8 & 33.4  & 31.4  & 64.9  & 60.8 \\
    Ours (NP)     &  28.7    & 26.0  & 54.0  & 33.9 & 40.9  & 34.1  & 68.7  & 79.7 \\
    Ours (NP+SG)    & 29.7     & 26.2  & 54.8  & 37.3 & 45.0  & 35.6  & 71.1  & 89.2  \\
    Ours (NP + SG + EC)   & 31.4  & 26.5  & 55.1  & 39.3& 45.7  & 35.9  & 71.5  & 92.9 \\
    \bottomrule
    \end{tabular}}
    \caption{Ablation study on multi-granularity semantic prompts. NP: Noun Phrase, SG: Scene Graph, EC: Entire Caption.}
  \label{tab:ablation_multiGrained}
\end{table}%

% \begin{figure}
%     \centering
%     \includegraphics[width=1\linewidth]{ICCV2025-Author-Kit/Figures/cross_domain_self_bleu.svg.pdf}
%     \caption{Self-BLEU to demonstrate diversity of noun phrases generated by our retrieval approach.}
%     \label{fig:Self-BLEU}
% \end{figure}

\begin{table}[t]
\centering
\resizebox{\linewidth}{!}{
\begin{tabular}{cccccc}
\toprule
Method & B@4 & M & R & C & Self-BLEU ($\downarrow$) \\
\midrule
\multicolumn{6}{c}{MSVD $\Rightarrow$ MSR-VTT} \\
\hline
Direct Top-\textit{K} & 19.7 & 21.7 & 50.7 & 20.9 & 0.565 \\
Category-aware & 19.9 & 22.1 & 51.3 & 19.1 & 0.432 \\
Category-aware w/ Top-\textit{p} & 20.9 & 22.8 & 51.9 & 23.2 & 0.419 \\
\midrule
\multicolumn{6}{c}{MSR-VTT $\Rightarrow$ MSVD} \\
\hline
Direct Top-\textit{K}  & 21.3 & 28.2 & 55.9 & 34.7 & 0.598 \\
Category-aware  & 23.0 & 28.0 & 57.6 & 31.0 & 0.538 \\
Category-aware w/ Top-\textit{p} & 25.2 & 29.1 & 58.7 & 41.4 & 0.475 \\
\bottomrule
\end{tabular}}
\caption{Ablation study on retrieval strategies, with metrics for caption quality and prompt diversity.
% Methods include \textbf{Direct Top-K Retrieval} (selecting the top-30 noun phrases by similarity), \textbf{Category-based Retrieval} (selecting 30 noun phrases across categories), and \textbf{Category-based Retrieval with Top-\textit{p} Filtering} (applying Top-\textit{p} post-processing). 
% NP: Noun Phrase. 
}
\label{tab:ablation_retrieval}
\end{table}

\begin{table}[tbp]
  \centering
  \resizebox{\linewidth}{!}{
    \begin{tabular}{cccccccccc}
    \toprule
    \multirow{2}{*}{Pre-trained VL Model} & \multirow{2}{*}{Size} & \multicolumn{4}{c}{VATEX} & \multicolumn{4}{c}{MSR-VTT} \\
    \cmidrule(lr){3-6} \cmidrule(lr){7-10}
    & & B@4 & M & R & C & B@4 & M & R & C \\
    \midrule
    CLIP (ViT/B-16) & 0.15B & 23.8 & 21.0 & 44.5 & 41.4 & 31.4 & 26.5 & 55.1 & 39.3 \\
    GME-Qwen2VL-2B & 2.2B & 26.3 & 23.2 & 46.3 & 50.9 & 31.8 & 26.6 & 56.1 & 41.5 \\
    GME-Qwen2VL-7B & 8.2B & 32.7 & 24.5 & 49.9 & 62.2 & 34.8 & 28.5 & 57.7 & 48.2 \\
    \bottomrule
    \end{tabular}}
    \caption{Impact of different scales of pre-trained vision-language models on in-domain zero-shot video captioning performance.}
  \label{tab:Pre_trained_VL_Scaling}%
\end{table}%

\subsection{Qualitative Analysis}\label{Qualitative}
% \paragraph{Comparison with Other Methods}

\begin{figure}
    
    \centering
    \includegraphics[width=1\linewidth]{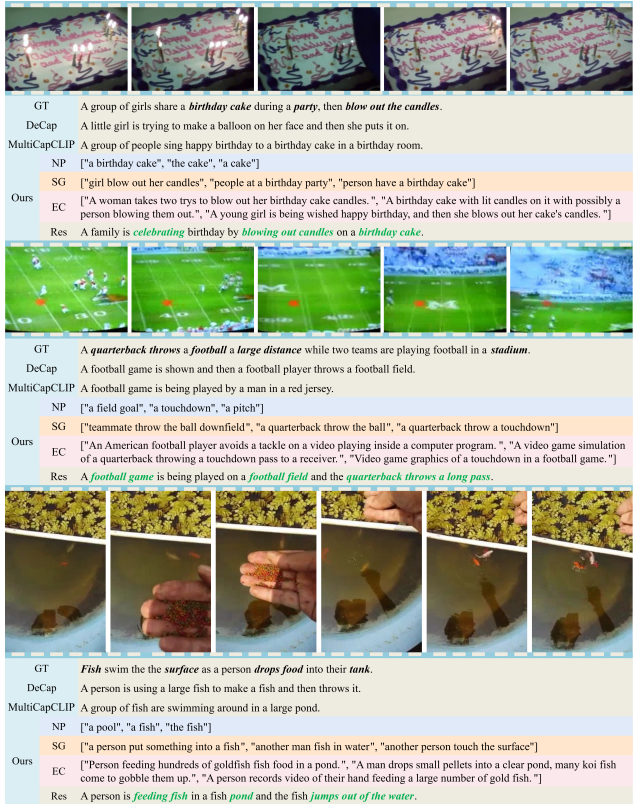}

    \caption{Comparison of generated captions of our method and other state-of-the-art methods. We emphasize ground-truth \textbf{\textit{important}} words and \textbf{\textit{\textcolor[RGB]{0,176,80}{accurate}}} words in our generated descriptions.}
    \label{fig:comparison}
\end{figure}

\cref{fig:comparison} provides a qualitative comparison of the generated captions from our method and other state-of-the-art approaches on three example videos. 
% For each video, we show the top 3 noun phrases (NP), 3 scene graphs, and 3 entire captions retrieved by our model. The ground-truth important words are highlighted in \textbf{black}, while the accurate words in the generated captions are marked in \textbf{green}. 
In the first video, our model precisely identifies the key entity ``birthday cake'' from the noun phrase memory bank, and, by leveraging both the scene graph and entire caption prompts, recognizes the action ``blow out candles'' which enables the model to generate a caption that covers all the essential information. For the remaining two videos, our three types of prompts continue to work synergistically. 
% The scene graph and entire caption provide complementary information, with the scene graph focusing on action details and the entire caption capturing the global context of the event. 
Through our designed retrieval method, noun phrases and scene graphs provide fundamental entities and action details, respectively, while entire captions capture the global context.
As a result, our method produces more comprehensive captions.
% compared to other approaches.
% , which often fail to capture the full complexity of the scenes. 
% This demonstrates that our model retrieves highly semantic-related multi-scale units that effectively guide caption generation.

\section{Conclusion}
We propose a novel zero-shot video captioning framework featuring two core innovations. First, our progressive multi-granularity prompting strategy hierarchically combines noun phrases (capturing fine-grained entities), attribute-enriched scene graphs (modeling structured object interactions), and entire captions (preserving contextual coherence) to comprehensively represent visual semantics. Second, we introduce category-aware retrieval with top-\textit{p} filtering, which leverages statistical priors from the training data for adaptive selection to ensure prompt diversity, and employs top-\textit{p} filtering to maintain semantic relevance.
Experiments on MSR-VTT, MSVD, and VATEX achieve new SoTA results. Ablation studies confirm our design's effectiveness and highlight potential for further improvement with future larger pre-trained vision-language models.

% Experiments on MSR-VTT, MSVD, and VATEX achieve new SoTA performance.
% % under in-domain and cross-domain settings. 
% Ablations not only confirm our design’s effectiveness, 
% % , while qualitative results demonstrate improved caption informativeness. 
% but also demonstrate potential for further improvement by utilizing larger pre-trained vision-language models in the future.
% suggesting broader applicability.
{
    \small
    \bibliographystyle{ieeenat_fullname}
    \bibliography{main}

\begin{thebibliography}{45}
\providecommand{\natexlab}[1]{#1}
\providecommand{\url}[1]{\texttt{#1}}
\expandafter\ifx\csname urlstyle\endcsname\relax
  \providecommand{\doi}[1]{doi: #1}\else
  \providecommand{\doi}{doi: \begingroup \urlstyle{rm}\Url}\fi

\bibitem[Banerjee and Lavie(2005)]{DBLP:conf/acl/METEOR}
Satanjeev Banerjee and Alon Lavie.
\newblock {METEOR:} an automatic metric for {MT} evaluation with improved correlation with human judgments.
\newblock In \emph{IEEvaluation@ACL}, pages 65--72, 2005.

\bibitem[Carreira and Zisserman(2017)]{DBLP:conf/cvpr/I3D}
Jo{\~{a}}o Carreira and Andrew Zisserman.
\newblock Quo vadis, action recognition? {A} new model and the kinetics dataset.
\newblock In \emph{{CVPR}}, pages 4724--4733, 2017.

\bibitem[Chen and Dolan(2011)]{DBLP:conf/acl/MSVD}
David~L. Chen and William~B. Dolan.
\newblock Collecting highly parallel data for paraphrase evaluation.
\newblock In \emph{{ACL}}, pages 190--200, 2011.

\bibitem[Cornia et~al.(2020)Cornia, Stefanini, Baraldi, and Cucchiara]{DBLP:conf/cvpr/M²_Transformer}
Marcella Cornia, Matteo Stefanini, Lorenzo Baraldi, and Rita Cucchiara.
\newblock Meshed-memory transformer for image captioning.
\newblock In \emph{{CVPR}}, pages 10575--10584, 2020.

\bibitem[Devlin et~al.(2019)Devlin, Chang, Lee, and Toutanova]{DBLP:conf/naacl/BERT}
Jacob Devlin, Ming{-}Wei Chang, Kenton Lee, and Kristina Toutanova.
\newblock {BERT:} pre-training of deep bidirectional transformers for language understanding.
\newblock In \emph{{NAACL-HLT}}, pages 4171--4186, 2019.

\bibitem[Fei et~al.(2023)Fei, Wang, Zhang, He, Wang, and Zheng]{DBLP:conf/iccv/ViECap}
Junjie Fei, Teng Wang, Jinrui Zhang, Zhenyu He, Chengjie Wang, and Feng Zheng.
\newblock Transferable decoding with visual entities for zero-shot image captioning.
\newblock In \emph{{ICCV}}, pages 3113--3123, 2023.

\bibitem[Gao et~al.(2022)Gao, Lei, Zeng, Song, Wang, and Shen]{DBLP:journals/tip/HRNAT}
Lianli Gao, Yu Lei, Pengpeng Zeng, Jingkuan Song, Meng Wang, and Heng~Tao Shen.
\newblock Hierarchical representation network with auxiliary tasks for video captioning and video question answering.
\newblock \emph{{IEEE} Trans. Image Process.}, 31:\penalty0 202--215, 2022.

\bibitem[Hara et~al.(2018)Hara, Kataoka, and Satoh]{DBLP:conf/cvpr/C3D}
Kensho Hara, Hirokatsu Kataoka, and Yutaka Satoh.
\newblock Can spatiotemporal 3d cnns retrace the history of 2d cnns and imagenet?
\newblock In \emph{{CVPR}}, pages 6546--6555, 2018.

\bibitem[He et~al.(2016)He, Zhang, Ren, and Sun]{DBLP:conf/cvpr/ResNet}
Kaiming He, Xiangyu Zhang, Shaoqing Ren, and Jian Sun.
\newblock Deep residual learning for image recognition.
\newblock In \emph{{CVPR}}, pages 770--778, 2016.

\bibitem[He(2021)]{DBLP:conf/emnlp/CBART}
Xingwei He.
\newblock Parallel refinements for lexically constrained text generation with {BART}.
\newblock In \emph{{EMNLP}}, pages 8653--8666, 2021.

\bibitem[Hochreiter and Schmidhuber(1997)]{DBLP:journals/neco/HochreiterS97}
Sepp Hochreiter and J{\"{u}}rgen Schmidhuber.
\newblock Long short-term memory.
\newblock \emph{Neural Comput.}, 9\penalty0 (8):\penalty0 1735--1780, 1997.

\bibitem[Holtzman et~al.(2020)Holtzman, Buys, Du, Forbes, and Choi]{Top_p}
Ari Holtzman, Jan Buys, Li Du, Maxwell Forbes, and Yejin Choi.
\newblock The curious case of neural text degeneration.
\newblock In \emph{{ICLR}}, 2020.

\bibitem[Jia et~al.(2022)Jia, Tang, Chen, Cardie, Belongie, Hariharan, and Lim]{VPT}
Menglin Jia, Luming Tang, Bor{-}Chun Chen, Claire Cardie, Serge~J. Belongie, Bharath Hariharan, and Ser{-}Nam Lim.
\newblock Visual prompt tuning.
\newblock In \emph{{ECCV}}, pages 709--727, 2022.

\bibitem[Krishna et~al.(2017)Krishna, Zhu, Groth, Johnson, Hata, Kravitz, Chen, Kalantidis, Li, Shamma, Bernstein, and Fei{-}Fei]{DBLP:journals/ijcv/Visual_Genome}
Ranjay Krishna, Yuke Zhu, Oliver Groth, Justin Johnson, Kenji Hata, Joshua Kravitz, Stephanie Chen, Yannis Kalantidis, Li{-}Jia Li, David~A. Shamma, Michael~S. Bernstein, and Li Fei{-}Fei.
\newblock Visual genome: Connecting language and vision using crowdsourced dense image annotations.
\newblock \emph{Int. J. Comput. Vis.}, 123\penalty0 (1):\penalty0 32--73, 2017.

\bibitem[Li et~al.(2024)Li, Vo, Sugimoto, and Nakayama]{DBLP:conf/cvpr/Evcap}
Jiaxuan Li, Duc~Minh Vo, Akihiro Sugimoto, and Hideki Nakayama.
\newblock Evcap: Retrieval-augmented image captioning with external visual-name memory for open-world comprehension.
\newblock In \emph{{CVPR}}, pages 13733--13742, 2024.

\bibitem[Li et~al.(2023{\natexlab{a}})Li, Zhu, Wen, and Yang]{DBLP:conf/iclr/DeCap}
Wei Li, Linchao Zhu, Longyin Wen, and Yi Yang.
\newblock Decap: Decoding {CLIP} latents for zero-shot captioning via text-only training.
\newblock In \emph{{ICLR}}, 2023{\natexlab{a}}.

\bibitem[Li et~al.(2023{\natexlab{b}})Li, Chai, Zhuo, Qu, Haffari, Li, Ji, and Tran]{DBLP:conf/acl/FACTUAL}
Zhuang Li, Yuyang Chai, Terry~Yue Zhuo, Lizhen Qu, Gholamreza Haffari, Fei Li, Donghong Ji, and Quan~Hung Tran.
\newblock {FACTUAL:} {A} benchmark for faithful and consistent textual scene graph parsing.
\newblock In \emph{{ACL} (Findings)}, pages 6377--6390, 2023{\natexlab{b}}.

\bibitem[Lin and Och(2004)]{DBLP:conf/acl/ROUGE-L}
Chin{-}Yew Lin and Franz~Josef Och.
\newblock Automatic evaluation of machine translation quality using longest common subsequence and skip-bigram statistics.
\newblock In \emph{{ACL}}, pages 605--612, 2004.

\bibitem[Lin et~al.(2022)Lin, Li, Lin, Ahmed, Gan, Liu, Lu, and Wang]{DBLP:conf/cvpr/swinbert}
Kevin Lin, Linjie Li, Chung{-}Ching Lin, Faisal Ahmed, Zhe Gan, Zicheng Liu, Yumao Lu, and Lijuan Wang.
\newblock Swinbert: End-to-end transformers with sparse attention for video captioning.
\newblock In \emph{{CVPR}}, pages 17928--17937, 2022.

\bibitem[Liu et~al.(2023)Liu, Wang, Zhang, Zheng, Jiang, and Lu]{STR}
Zhu Liu, Teng Wang, Jinrui Zhang, Feng Zheng, Wenhao Jiang, and Ke Lu.
\newblock Show, tell and rephrase: Diverse video captioning via two-stage progressive training.
\newblock \emph{{IEEE} Trans. Multim.}, 25:\penalty0 7894--7905, 2023.

\bibitem[Loshchilov and Hutter(2019)]{AdamW}
Ilya Loshchilov and Frank Hutter.
\newblock Decoupled weight decay regularization.
\newblock In \emph{{ICLR}}, 2019.

\bibitem[Nukrai et~al.(2022)Nukrai, Mokady, and Globerson]{DBLP:conf/emnlp/CapDec}
David Nukrai, Ron Mokady, and Amir Globerson.
\newblock Text-only training for image captioning using noise-injected {CLIP}.
\newblock In \emph{{EMNLP} (Findings)}, pages 4055--4063, 2022.

\bibitem[OpenAI(2023)]{gpt4}
OpenAI.
\newblock {GPT-4} technical report.
\newblock \emph{CoRR}, abs/2303.08774, 2023.

\bibitem[Papineni et~al.(2002)Papineni, Roukos, Ward, and Zhu]{DBLP:conf/acl/BLEU4}
Kishore Papineni, Salim Roukos, Todd Ward, and Wei{-}Jing Zhu.
\newblock Bleu: a method for automatic evaluation of machine translation.
\newblock In \emph{{ACL}}, pages 311--318, 2002.

\bibitem[Radford et~al.(2019)Radford, Wu, Child, Luan, Amodei, Sutskever, et~al.]{GPT_2}
Alec Radford, Jeffrey Wu, Rewon Child, David Luan, Dario Amodei, Ilya Sutskever, et~al.
\newblock Language models are unsupervised multitask learners.
\newblock \emph{OpenAI blog}, 1\penalty0 (8):\penalty0 9, 2019.

\bibitem[Radford et~al.(2021)Radford, Kim, Hallacy, Ramesh, Goh, Agarwal, Sastry, Askell, Mishkin, Clark, Krueger, and Sutskever]{DBLP:conf/icml/CLIP}
Alec Radford, Jong~Wook Kim, Chris Hallacy, Aditya Ramesh, Gabriel Goh, Sandhini Agarwal, Girish Sastry, Amanda Askell, Pamela Mishkin, Jack Clark, Gretchen Krueger, and Ilya Sutskever.
\newblock Learning transferable visual models from natural language supervision.
\newblock In \emph{{ICML}}, pages 8748--8763, 2021.

\bibitem[Ryu et~al.(2021)Ryu, Kang, Kang, and Yoo]{DBLP:conf/aaai/SGN}
Hobin Ryu, Sunghun Kang, Haeyong Kang, and Chang~D. Yoo.
\newblock Semantic grouping network for video captioning.
\newblock In \emph{{AAAI}}, pages 2514--2522, 2021.

\bibitem[Shen et~al.(2023)Shen, Gu, Xu, Fan, Wen, and Zhang]{CoCap}
Yaojie Shen, Xin Gu, Kai Xu, Heng Fan, Longyin Wen, and Libo Zhang.
\newblock Accurate and fast compressed video captioning.
\newblock In \emph{{ICCV}}, pages 15558--15567, 2023.

\bibitem[Su et~al.(2022)Su, Lan, Liu, Liu, Yogatama, Wang, Kong, and Collier]{DBLP:journals/corr/MAGIC}
Yixuan Su, Tian Lan, Yahui Liu, Fangyu Liu, Dani Yogatama, Yan Wang, Lingpeng Kong, and Nigel Collier.
\newblock Language models can see: Plugging visual controls in text generation.
\newblock \emph{CoRR}, abs/2205.02655, 2022.

\bibitem[Tewel et~al.(2022)Tewel, Shalev, Schwartz, and Wolf]{DBLP:conf/cvpr/ZeroCap}
Yoad Tewel, Yoav Shalev, Idan Schwartz, and Lior Wolf.
\newblock Zerocap: Zero-shot image-to-text generation for visual-semantic arithmetic.
\newblock In \emph{{CVPR}}, pages 17897--17907. {IEEE}, 2022.

\bibitem[Tewel et~al.(2023)Tewel, Shalev, Nadler, Schwartz, and Wolf]{DBLP:conf/bmvc/ZeroCap_video}
Yoad Tewel, Yoav Shalev, Roy Nadler, Idan Schwartz, and Lior Wolf.
\newblock Zero-shot video captioning by evolving pseudo-tokens.
\newblock In \emph{{BMVC}}, pages 429--432, 2023.

\bibitem[Tian et~al.(2024)Tian, Li, Qi, Wang, Sheng, and Huang]{DBLP:journals/pr/tmk_paper}
Mingkai Tian, Guorong Li, Yuankai Qi, Shuhui Wang, Quan~Z. Sheng, and Qingming Huang.
\newblock Rethink video retrieval representation for video captioning.
\newblock \emph{Pattern Recognit.}, 156:\penalty0 110744, 2024.

\bibitem[Vaswani et~al.(2017)Vaswani, Shazeer, Parmar, Uszkoreit, Jones, Gomez, Kaiser, and Polosukhin]{vanilla_transformer}
Ashish Vaswani, Noam Shazeer, Niki Parmar, Jakob Uszkoreit, Llion Jones, Aidan~N. Gomez, Lukasz Kaiser, and Illia Polosukhin.
\newblock Attention is all you need.
\newblock In \emph{{NIPS}}, pages 5998--6008, 2017.

\bibitem[Vedantam et~al.(2015)Vedantam, Zitnick, and Parikh]{DBLP:conf/cvpr/CIDEr}
Ramakrishna Vedantam, C.~Lawrence Zitnick, and Devi Parikh.
\newblock Cider: Consensus-based image description evaluation.
\newblock In \emph{{CVPR}}, pages 4566--4575, 2015.

\bibitem[Wang et~al.(2019{\natexlab{a}})Wang, Ma, Zhang, Jiang, Wang, and Liu]{DBLP:conf/iccv/POS-CG}
Bairui Wang, Lin Ma, Wei Zhang, Wenhao Jiang, Jingwen Wang, and Wei Liu.
\newblock Controllable video captioning with {POS} sequence guidance based on gated fusion network.
\newblock In \emph{{ICCV}}, pages 2641--2650, 2019{\natexlab{a}}.

\bibitem[Wang et~al.(2019{\natexlab{b}})Wang, Wu, Chen, Li, Wang, and Wang]{VATEX}
Xin Wang, Jiawei Wu, Junkun Chen, Lei Li, Yuan{-}Fang Wang, and William~Yang Wang.
\newblock Vatex: {A} large-scale, high-quality multilingual dataset for video-and-language research.
\newblock In \emph{{ICCV}}, pages 4580--4590, 2019{\natexlab{b}}.

\bibitem[Xiao et~al.(2024)Xiao, Liu, Zhang, Muennighoff, Lian, and Nie]{BGE}
Shitao Xiao, Zheng Liu, Peitian Zhang, Niklas Muennighoff, Defu Lian, and Jian-Yun Nie.
\newblock C-pack: Packed resources for general chinese embeddings.
\newblock In \emph{{SIGIR}}, pages 641–--649, 2024.

\bibitem[Xu et~al.(2016)Xu, Mei, Yao, and Rui]{DBLP:conf/cvpr/MSR-VTT}
Jun Xu, Tao Mei, Ting Yao, and Yong Rui.
\newblock {MSR-VTT:} {A} large video description dataset for bridging video and language.
\newblock In \emph{{CVPR}}, pages 5288--5296, 2016.

\bibitem[Yan et~al.(2025)Yan, Xie, Zou, Wei, and Luan]{DBLP:journals/ijon/EntroCap}
Jie Yan, Yuxiang Xie, Shiwei Zou, Yingmei Wei, and Xidao Luan.
\newblock Entrocap: Zero-shot image captioning with entropy-based retrieval.
\newblock \emph{Neurocomputing}, 611:\penalty0 128666, 2025.

\bibitem[Yang et~al.(2023)Yang, Liu, Wu, Wang, Sun, and Zou]{DBLP:conf/acl/MultiCapCLIP}
Bang Yang, Fenglin Liu, Xian Wu, Yaowei Wang, Xu Sun, and Yuexian Zou.
\newblock Multicapclip: Auto-encoding prompts for zero-shot multilingual visual captioning.
\newblock In \emph{{ACL}}, pages 11908--11922, 2023.

\bibitem[Zeng et~al.(2023)Zeng, Zhang, Lu, Wang, Chen, and Wang]{DBLP:conf/cvpr/ConZIC}
Zequn Zeng, Hao Zhang, Ruiying Lu, Dongsheng Wang, Bo Chen, and Zhengjue Wang.
\newblock Conzic: Controllable zero-shot image captioning by sampling-based polishing.
\newblock In \emph{{CVPR}}, pages 23465--23476, 2023.

\bibitem[Zeng et~al.(2024)Zeng, Xie, Zhang, Chen, Chen, and Wang]{DBLP:conf/cvpr/MeaCap}
Zequn Zeng, Yan Xie, Hao Zhang, Chiyu Chen, Bo Chen, and Zhengjue Wang.
\newblock Meacap: Memory-augmented zero-shot image captioning.
\newblock In \emph{{CVPR}}, pages 14100--14110, 2024.

\bibitem[Zhang et~al.(2024)Zhang, Zhang, Xie, Li, Dai, Long, Xie, Zhang, Li, and Zhang]{DBLP:journals/corr/GME}
Xin Zhang, Yanzhao Zhang, Wen Xie, Mingxin Li, Ziqi Dai, Dingkun Long, Pengjun Xie, Meishan Zhang, Wenjie Li, and Min Zhang.
\newblock {GME:} improving universal multimodal retrieval by multimodal llms.
\newblock \emph{CoRR}, abs/2412.16855, 2024.

\bibitem[Zheng et~al.(2020)Zheng, Wang, and Tao]{DBLP:conf/cvpr/SAAT}
Qi Zheng, Chaoyue Wang, and Dacheng Tao.
\newblock Syntax-aware action targeting for video captioning.
\newblock In \emph{{CVPR}}, pages 13093--13102, 2020.

\bibitem[Zhu et~al.(2018)Zhu, Lu, Zheng, Guo, Zhang, Wang, and Yu]{self_BLEU}
Yaoming Zhu, Sidi Lu, Lei Zheng, Jiaxian Guo, Weinan Zhang, Jun Wang, and Yong Yu.
\newblock Texygen: {A} benchmarking platform for text generation models.
\newblock In \emph{{SIGIR}}, pages 1097--1100, 2018.

\end{thebibliography}
}

\clearpage
\setcounter{page}{1}
\maketitlesupplementary

\section{Implementation Details}\label{suppleDetail}
This section provides additional details on our implementation.
% , as described in~\cref{Experimental_Setups} of the main paper
We adopt the same model architecture as MultiCapCLIP~\cite{DBLP:conf/acl/MultiCapCLIP}, where the text decoder consists of a 6-layer Transformer~\cite{vanilla_transformer} with 8 attention heads and a hidden size of 512. The CLIP (ViT/B-16)~\cite{DBLP:conf/icml/CLIP} model encodes textual units, which are subsequently processed by a feed-forward network (FFN) with both input and output dimensions set to 512 before being fed into the text decoder. During training, we apply label smoothing with a value of 0.1, while for inference, we employ beam search with a beam size of 3 to generate text tokens. All experiments are conducted over 10 epochs using the AdamW~\cite{AdamW} optimizer, incorporating a linear warm-up phase over the first 10\% of the training steps.

The threshold \(\tau\) for top-\(p\) post-processing remains consistent across noun phrases and scene graphs. For in-domain tasks, the MSR-VTT~\cite{DBLP:conf/cvpr/MSR-VTT} and MSVD~\cite{DBLP:conf/acl/MSVD} datasets utilize a peak learning rate of \(1 \times 10^{-4}\), which remains fixed after the warm-up phase. In contrast, the VATEX~\cite{VATEX} dataset employs a peak learning rate of \(5 \times 10^{-4}\), followed by a linear decay to 0 after the warm-up period. The value of \(\tau\) is set to 0.6 for both MSVD and VATEX, and to 0.8 for MSR-VTT.

In the cross-domain setting, for the MSR-VTT \(\Rightarrow\) MSVD task, the parameters \(K_p\) and \(K_g\) are configured to 12 and 34, respectively. For the MSVD \(\Rightarrow\) MSR-VTT task, these parameters are set to 14 and 25, respectively. The learning rate and scheduler configurations mirror those of the in-domain tasks, with \(\tau\) fixed at 0.5 for both cross-domain tasks.

\section{Scene Graph Memory Bank Construction}
\cref{alg:sg_memory_v2} details the process of constructing the enhanced scene graph memory bank.
% , as described in~\cref{subsec:memory_bank} of the main paper. 
It takes the training captions as input and generates a memory bank consisting of scene graphs enriched with noun phrases.
\begin{algorithm}[ht]
\small
\SetAlgoLined
\KwIn{$\mathcal{S}$: Training captions\; $\{\mathcal{P}(S)\}_{S\in\mathcal{S}}$: Noun phrase sets\; $N_g$: Frequency threshold}
\KwOut{$\mathcal{M}_{\text{SG}}$: Enhanced scene graph memory bank}
% \BlankLine
$\mathcal{X}_{\text{all}} \gets \varnothing$\;
\For{caption $S \in \mathcal{S}$}{
    $\mathcal{G}_S \gets \text{TextualParser}(S)$\;
    % \tcp*{Parse $\langle sub,pred,obj \rangle$ triples} 
    \For{ $g_i = \langle sub_i, pred_i, obj_i \rangle \in \mathcal{G}_S$}{
        % \tcp{Step 1: Noun phrase augmentation for subject/object}
        $\mathcal{A}_i \gets \{p  \mid  p\in \mathcal{P}(S) \land sub_i \text{ is substring of } p\} \cup \{sub_i\}$\;
        $\mathcal{B}_i \gets \{p  \mid  p\in \mathcal{P}(S) \land obj_i \text{ is substring of } p\} \cup \{obj_i\}$\;
        
        % \tcp{Step 2: Generate candidate scene graphs}
        $\mathcal{X}_i \gets \{\langle a, pred_i, b \rangle \mid a \in \mathcal{A}_i, b \in \mathcal{B}_i\}$\;
        
        % \tcp{Step 3: Semantic alignment selection}
        $\mathbf{E}_S \gets \text{BGE}(S)$\;
        % \tcp*{Encode entire caption}
        $\mathbf{E}_{\mathcal{X}_i} \gets \text{BGE}(\mathcal{X}_i)$\;
        % \tcp*{Batch encode candidates}
        $x_{\text{best}} \gets \arg\max_{x_i^j\in\mathcal{X}_i} \cos(\mathbf{E}_S, \mathbf{E}_{\mathcal{X}_i}[x_i^j])$\;
        
        % \tcp{Collect best candidate per original graph}
        $\mathcal{X}_{\text{all}} \gets \mathcal{X}_{\text{all}} \cup \{x_{\text{best}}\}$\;
    }
}
% \BlankLine
% \tcp{Frequency-based memory bank construction}
$\mathcal{F} \gets \{(x, \text{count}(x \in \mathcal{X}_{\text{all}})) \mid x \in \mathcal{X}_{\text{all}}\}$\;
% \tcp*{Count occurrences}
$\mathcal{M}_{\text{SG}} \gets \text{Sort}(\mathcal{F}, \text{by count descending})[0:N_g-1]$\;
% \tcp*{Top-$N_2$ selection}
\Return $\mathcal{M}_{\text{SG}}$
\caption{\small Enhanced Scene Graph Memory Bank Construction}
\label{alg:sg_memory_v2}
\end{algorithm}

\section{Classification of Noun Phrase Memory Bank}\label{Classification_Result_of_NP_Memory_Bank}

\begin{figure*}
    \centering
    \includegraphics[width=1\linewidth]{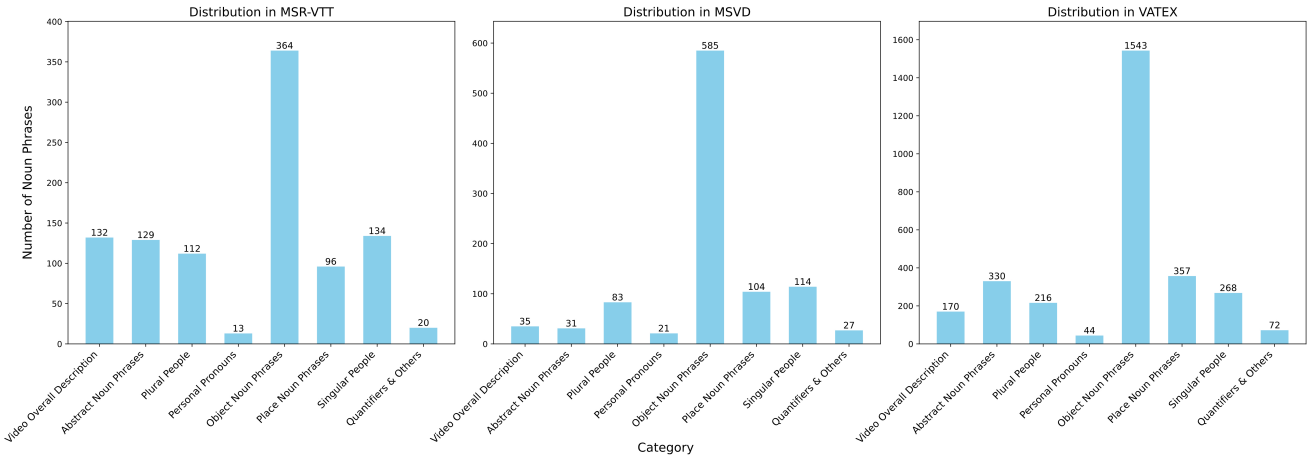}
    \caption{Distribution of Noun Phrases across Different Categories in the MSR-VTT, MSVD, and VATEX Datasets.}
    \label{fig:noun_phrase_distribution}
\end{figure*}

% Expanding on the classification process described in~\cref{Noun Phrase Prompt Generation}, 
We first performed unsupervised classification using GPT-4~\cite{gpt4} on the noun phrase memory bank \( \mathcal{M}_{\text{NP}} \) of MSR-VTT. Subsequently, the same categories are applied to the classification process for the MSVD and VATEX datasets. \cref{tab:gpt_categories} presents the eight categories identified by GPT-4, together with their interpretations. Examples of noun phrases belonging to each category are provided in the last column. \cref{fig:noun_phrase_distribution} illustrates the distribution of noun phrases across these categories in the MSR-VTT, MSVD, and VATEX datasets. As can be observed, object noun phrases consistently dominate across all datasets, followed by singular people. For the complex VATEX dataset, which contains more diverse and intricate scenes, place noun phrases also exhibit a significant presence.

\begin{table*}[!]
\centering
\small % Reduce font size for better fit
\resizebox{\linewidth}{!}{
\begin{tabular}{cccc}
\toprule
Category ID & Category Name & Category Description & Example Noun Phrases \\
\midrule
1 & Video Overall Description & Video types and content descriptions & ``a talk show", ``sports highlights", ``a music video" \\
2 & Abstract Noun Phrases & Noun phrases representing abstract concepts & ``different types", ``the features", ``beauty" \\
3 & Plural People & Nouns representing plural people & ``some kids", ``two men", ``a band" \\
4 & Personal Pronouns & Pronouns referring to people & ``he", ``us", ``they" \\
5 & Object Noun Phrases & Noun phrases representing physical objects & ``black t-shirt", ``a frying pan", ``a race car" \\
6 & Place Noun Phrases & Noun phrases representing different environments & ``a busy street", ``a basketball court", ``a restaurant" \\
7 & Singular People & Noun phrases representing single people & ``the man", ``a police officer", ``Captain America" \\
8 & Quantifiers \& Others & Quantifiers and other general terms & ``one", ``which", ``another" \\
\bottomrule
\end{tabular}
}
\caption{GPT-4-generated Categories for Noun Phrases Memory Bank \( \mathcal{M}_{\text{NP}} \) with Example Phrases.}
\label{tab:gpt_categories}
\end{table*}

\section{Other Ablation Studies}

\subsection{Impact of Memory Bank Size}

As illustrated in~\cref{tab:bank_size_full}, we evaluate the performance of in-domain zero-shot video captioning with varying sizes of memory bank of noun phrase and scene graph containing noun phrases. Following DeCap~\cite{DBLP:conf/iclr/DeCap}, where the prompt at the entire caption granularity occupies only one token, computations during retrieval are considerably simplified. Therefore, we fix the size of the entire caption memory bank to the number of captions in the training set. Upon increasing the sizes of both the noun phrase and scene graph memory bank, we observe improvements across all metrics, with the enhancement in CIDEr~\cite{DBLP:conf/cvpr/CIDEr} being the most notable. A larger memory bank suggests a more comprehensive and enriched knowledge base from the training set, thereby enhancing the generalization capability from training to inference phases.

\begin{table}[!]
  \centering
  \resizebox{0.85\linewidth}{!}{ % 全宽度自适应
    \begin{tabular}{cccccc}
    \toprule
    \multicolumn{2}{c}{Size of Memory Bank} & \multicolumn{4}{c}{VATEX} \\
    \cmidrule(lr){1-2} \cmidrule(lr){3-6}
    NP & SG & B@4 & M & R & C \\
    \midrule
    1000 & 200,000 & 23.1 & 20.6 & 44.1 & 37.4 \\
    1000 & 400,000 & 23.3 & 21.0 & 44.2 & 39.1 \\
    3000 & 400,000 & 23.8 & 21.0 & 44.5 & 41.4 \\
    \bottomrule
    \end{tabular}}
  % \vspace{-5pt}
  % \parbox{\textwidth}{\footnotesize \textbf{Bold} indicates best performance. CIDEr (C) improvements: +3.7\% (200k→400k SG) and +5.1\% (1000→3000 NP).
  \caption{Impact of memory bank size on in-domain captioning, evaluated on VATEX test set. NP: Noun Phrase, SG: Scene Graph.}
  \label{tab:bank_size_full}
\end{table}

\subsection{Impact of top-\textit{K} Selection from Memory}

\begin{figure}
    \centering
    \includegraphics[width=1\linewidth]{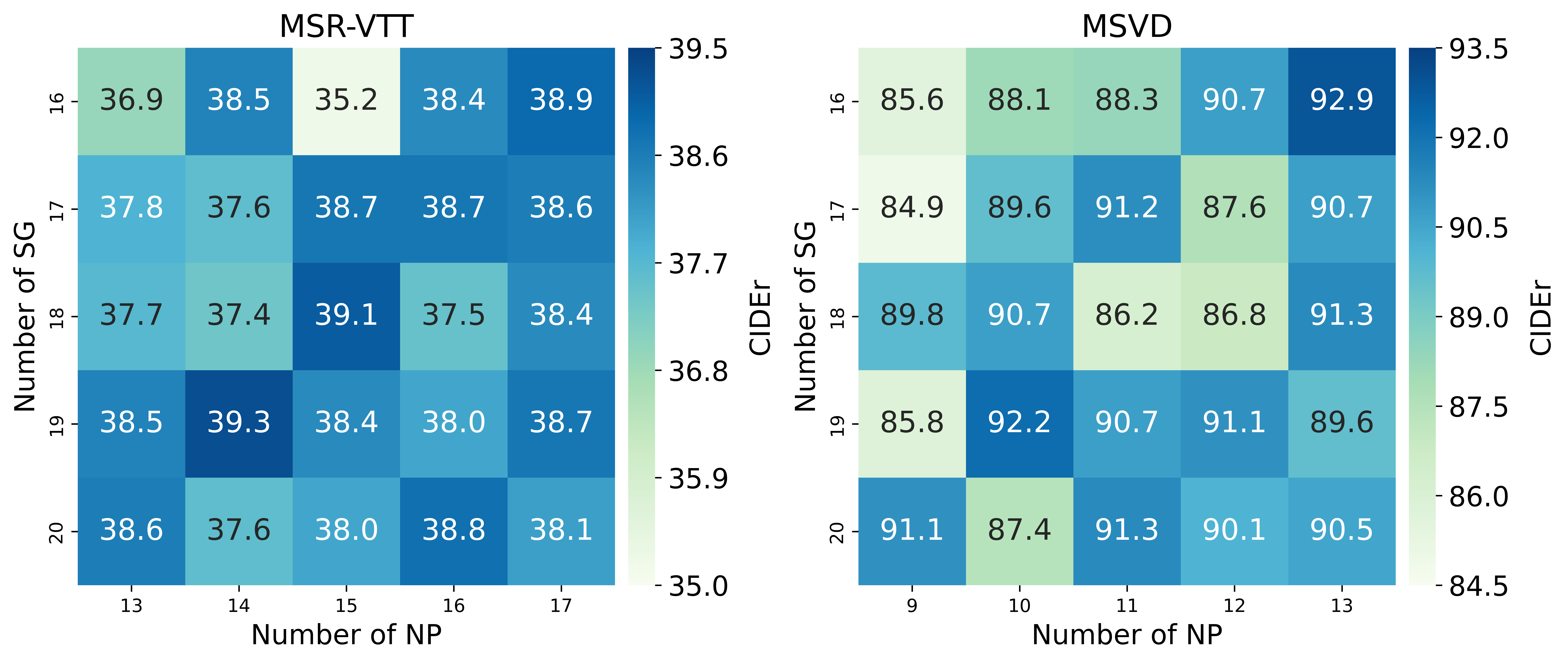}
    \caption{Impact of number of selected elements from noun phrase and scene graph memory banks on in-domain CIDEr scores. NP: Noun Phrase, SG: Scene Graph.}
    \label{fig:training_topk_heatmap}
\end{figure}

During training, we retrieve fixed numbers of elements from both the noun phrase (NP) and scene graph (SG) memory banks for each caption, which are then fed into the language decoder as prefix prompts for reconstruction. As visualized in~\cref{fig:training_topk_heatmap}, our ablation study on MSR-VTT and MSVD datasets systematically investigates how varying selection quantities of NP and SG elements affect the CIDEr metric. Notably, the model demonstrates well robustness across different parameter combinations, maintaining consistently high performance levels. 
% This stability suggests that our memory retrieval mechanism effectively captures semantically critical features regardless of selection thresholds. 
Through grid search optimization, we ultimately identify top-14 NP with top-19 SG as the optimal configuration for MSR-VTT, while top-13 NP paired with top-16 SG achieves peak performance on MSVD.

\subsection{Qualitative Gains from Scaling Multimodal Models}\label{scaleQuali}
% Building on the quantitative analysis of scaling up pre-trained multimodal models in~\cref{ablation} of the main paper, 
Building on the quantitative analysis of scaling up pre-trained multimodal models of the main paper, 
we present a qualitative analysis in~\cref{fig:scaleUp_quali}, visually illustrating the benefits. Compared to CLIP (ViT/B-16), larger models like GME-Qwen2VL-2B~\cite{DBLP:journals/corr/GME} and GME-Qwen2VL-7B~\cite{DBLP:journals/corr/GME} retrieve more video-relevant textual units from the same memory bank, leading to more semantically accurate and detail-rich captions.

\begin{figure}[!]
    \centering
    \includegraphics[width=1\linewidth]{Figures/quali_supplementary2.svg.pdf}
    \caption{Comparison of the three granularities of text prompts retrieved using different pre-trained multimodal models and the generated captions, denoted as 
    % \colorbox[RGB]{219,238,244}{\textcolor{black}{Res}}.
    ``Res''.
    We emphasize ground-truth \textbf{\textit{important}} words and \textbf{\textit{\textcolor[RGB]{0,176,80}{accurate}}} words in our generated descriptions.}
    \label{fig:scaleUp_quali}
\end{figure}

\end{document}